\documentclass[runningheads]{llncs}
\usepackage{graphicx}

\usepackage{tikz}
\usepackage{comment}
\usepackage{amsmath,amssymb} 
\usepackage{color}

\usepackage[accsupp]{axessibility}  

\usepackage[width=122mm,left=12mm,paperwidth=146mm,height=193mm,top=12mm,paperheight=217mm]{geometry}

\usepackage{booktabs}
\usepackage[pagebackref=true,breaklinks=true,colorlinks,bookmarks=false]{hyperref}
\usepackage{multirow}
\usepackage{diagbox}

\usepackage[caption=false]{subfig}
\usepackage{microtype}
\usepackage{wrapfig}

\usepackage{soul}
\renewcommand{\paragraph}[1]{\vspace{1mm}\noindent\textbf{#1}}
\newcommand{\mytabular}[1]{\scalebox{0.65}{#1}}
\newcommand{\secspaceA}{\vspace{-.35cm}}
\newcommand{\secspaceB}{\vspace{-.25cm}}
\newcommand{\subsecspaceA}{\vspace{-.21cm}}
\newcommand{\subsecspaceB}{\vspace{-.14cm}}

\usepackage[capitalize]{cleveref}
\crefname{section}{Sec.}{Secs.}
\Crefname{section}{Section}{Sections}
\Crefname{table}{Table}{Tables}
\crefname{table}{Table}{Tables}

\usepackage{xspace}
\makeatletter
\DeclareRobustCommand\onedot{\futurelet\@let@token\@onedot}
\def\@onedot{\ifx\@let@token.\else.\null\fi\xspace}

\def\eg{\emph{e.g}\onedot} \def\Eg{\emph{E.g}\onedot}
\def\ie{\emph{i.e}\onedot} \def\Ie{\emph{I.e}\onedot}
\def\cf{\emph{cf}\onedot} \def\Cf{\emph{Cf}\onedot}
\def\etc{\emph{etc}\onedot} \def\vs{\emph{vs}\onedot}
\def\wrt{w.r.t\onedot} \def\dof{d.o.f\onedot}
\def\iid{i.i.d\onedot} \def\wolog{w.l.o.g\onedot}
\def\etal{\emph{et al}\onedot}
\makeatother

\begin{document}
\pagestyle{headings}
\mainmatter
\def\ECCVSubNumber{4318}  

\title{One-Trimap Video Matting} 

\titlerunning{One-Trimap Video Matting}
%
\author{Hongje Seong\inst{1,}\thanks{This work was done during an internship at Adobe Research.} \and Seoung Wug Oh\inst{2} \and Brian Price\inst{2} \and \\ Euntai Kim\inst{1} \and Joon-Young Lee\inst{2}}
\authorrunning{H. Seong, S. W. Oh, B. Price, E. Kim, and J.-Y. Lee}
%
\institute{Yonsei University, Seoul, Korea. \email{\{hjseong,etkim\}@yonsei.ac.kr} \and Adobe Research, San Jose, CA, USA. \email{\{seoh,bprice,jolee\}@adobe.com}}
\maketitle

\begin{abstract}
Recent studies made great progress in video matting by extending the success of trimap-based image matting to the video domain.
In this paper, we push this task toward a more practical setting and propose One-Trimap Video Matting network (OTVM) that performs video matting robustly using only one user-annotated trimap.
A key of OTVM is the joint modeling of trimap propagation and alpha prediction.
Starting from baseline trimap propagation and alpha prediction networks, our OTVM combines the two networks with an alpha-trimap refinement module to facilitate information flow. 
We also present an end-to-end training strategy to take full advantage of the joint model.
Our joint modeling greatly improves the temporal stability of trimap propagation compared to the previous decoupled methods.
We evaluate our model on two latest video matting benchmarks, Deep Video Matting and VideoMatting108, and outperform state-of-the-art by significant margins (MSE improvements of 56.4\% and 56.7\%, respectively).
The source code and model are available online: \url{https://github.com/Hongje/OTVM}.
\end{abstract}

\secspaceA
\section{Introduction}
\label{sec:intro}
\secspaceB

\begin{figure}[t]
\centering
\includegraphics[width=.7\linewidth]{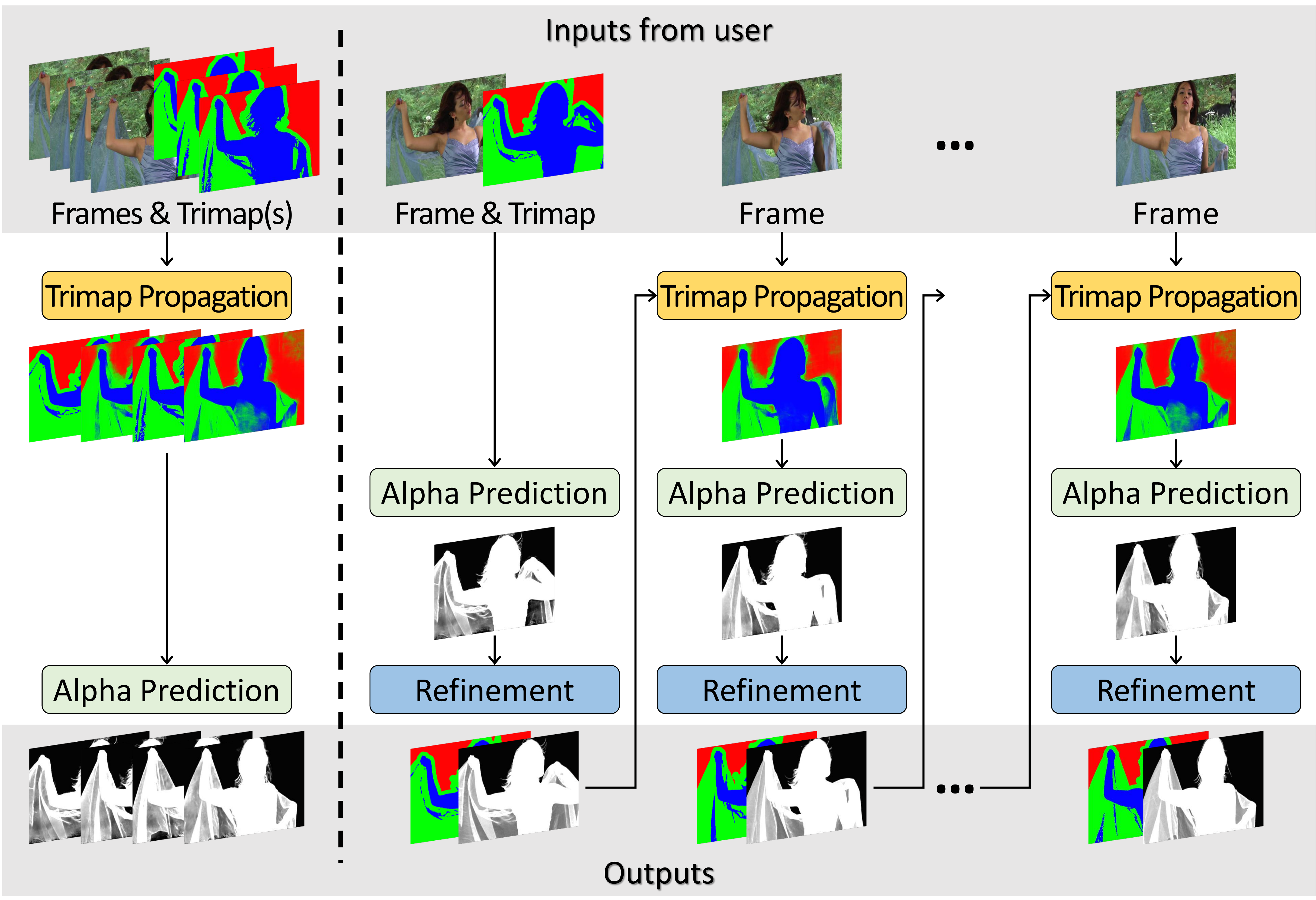}\\
\vspace{-0.1cm}
\raggedright{\scriptsize \hspace{45pt} (a) Previous~\cite{sun2021deep,zhang2021attention} \hspace{45pt} (b) OTVM (Ours)}
\vspace{-0.2cm}
\caption{
\textbf{The previous decoupled approach \textit{vs.} Our joint modeling}. (a) Previous video matting methods~\cite{sun2021deep,zhang2021attention} have two decoupled stages: first generate all missing trimaps and then predict the alpha mattes given the trimaps. This approach does not have any interaction between trimap propagation and alpha matting. Since the trimap propagation with no understanding of alpha mattes is often not stable, they require \textit{multiple user-annotated trimaps} to prevent drifting at trimap propagation.
(b) In our OTVM, trimap propagation and alpha prediction modules interact each other and the refinement step updates the predictions from the two modules. Our trimap propagation module memorizes and utilizes all information (RGB, trimap, alpha, latent features) to propagate trimaps. It results in accurate and robust predictions, enabling us to perform \textit{one-trimap video matting}.
\vspace{-0.5cm}
}
\label{fig:fig1}
\end{figure}

Video matting is the task of predicting accurate alpha mattes from a video.
This is an essential step in video editing applications requiring an accurate separation of the foreground and the background layers such as video composition. 
For each video frame $I$, it aims to divide the input color into three components: the foreground color, the background color, and the alpha matte. 
Formally, for a given pixel, it can be written as, $I = \alpha F + ( 1 - \alpha ) B,$ where $F$ and $B$ are the foreground and background color, and $\alpha \in [0, 1]$ represents the alpha value. 
Here, only 3 values ($I$) are known, and the remaining 7 values ($F$, $B$, and $\alpha$) are unknown.
Given the ill-posed nature of the problem, traditional methods utilize trimaps as additional inputs that indicate pixels that are either solid foreground, solid background, or uncertain.
The trimap provides a clue for the target object and effectively improves the stability of the alpha prediction.

Leveraging the latest progress in trimap-based image matting~\cite{xu2017deep,li2020natural,forte2020f} and mask propagation~\cite{xu2018ytvos,Oh_2019_ICCV,seong2021hierarchical}, recent studies~\cite{sun2021deep,zhang2021attention} successfully developed learning-based video matting techniques. They decouple video matting into two stages, trimap propagation and alpha prediction.
They re-purpose the latest mask propagation network~\cite{Oh_2019_ICCV} to propagate the given trimaps throughout the video, then design alpha prediction networks that take multiple trimaps as input and predict the alpha matte~(\cref{fig:fig1}(a)).

\begin{figure}[t]
\centering
\includegraphics[width=.8\linewidth]{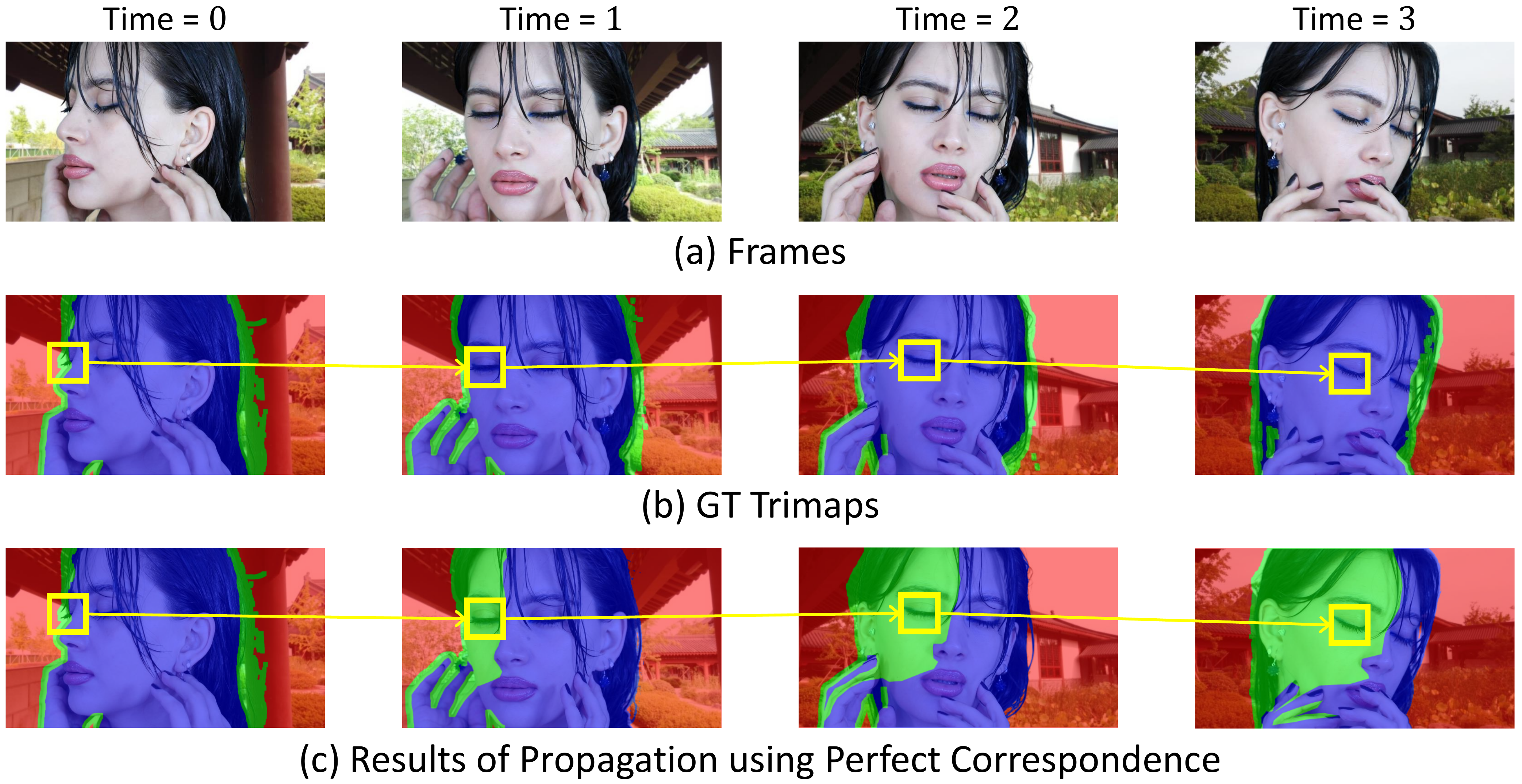}
\vspace{-0.4cm}
\caption{
\textbf{The intrinsic challenge of trimap propagation}.
As shown in (b), the same part of an instance can have varying trimap labels at different times, \eg, the label of the left eyelid has changed from the unknown to the foreground. If trimap propagation is conducted using visual correspondences without the knowledge for alpha matte, then it may produce inaccurate trimaps and the error can be easily accumulated as in (c).
\vspace{-0.5cm}
}
\label{fig:fig2}
\end{figure}

While the decoupled approach effectively simplifies the task, it has a critical limitation as illustrated in \cref{fig:fig2}. By the nature of the trimap, the unknown region of one frame may be changed into foreground or background and vice versa at a different frame. 
Therefore, if we propagate a trimap based on visual correspondences without the knowledge for alpha matte~\cite{Oh_2019_ICCV,cheng2021rethinking}, it may produce inaccurate trimaps and the error can be easily accumulated as shown in \cref{fig:fig2}(c), leading to the failure of alpha prediction. With this challenge, the existing decoupled methods require multiple user-annotated trimaps to prevent drifting at trimap propagation.

In this paper, we aim to tackle video matting with a single trimap input.
To cope with the challenging scenario, we propose One-Trimap Video Matting network (OTVM) that performs trimap propagation and alpha prediction as a joint task, as illustrated in \cref{fig:fig1}(b). 
Starting from baseline trimap propagation and alpha prediction modules, we cascade the two modules to alternate trimap propagation and alpha prediction auto-regressively at each frame. 
We employ the space-time memory (STM) network~\cite{Oh_2019_ICCV} and the FBA matting network~\cite{forte2020f} as the baseline modules, respectively. 
To facilitate information flow within the pipeline, we add a refinement module and re-engineer STM accordingly.
In addition, we present an end-to-end training pipeline to make OTVM learn the joint task successfully.
The major advantage of OTVM is robust trimap propagation that is critical for the practical video matting scenario.
Since an alpha matte contains richer information than a trimap, we are able to update the trimap after alpha prediction and this update step prevents error accumulation in the trimap, resulting in robust trimap propagation and accurate alpha prediction.

OTVM produces accurate video mattes even in the challenging one-trimap scenario.
We demonstrate that OTVM outperforms previous matting methods with large margins on two latest video matting benchmarks: 56.4\% improvement on Deep Video Matting (DVM) \cite{sun2021deep} and 56.7\% improvement on VideoMatting108 \cite{zhang2021attention} in terms of MSE. We also conduct extensive analysis experiments and show that the proposed joint modeling and learning scheme are crucial for achieving robust and accurate video matting results.

\secspaceA
\section{Related Work}
\secspaceB
\label{sec:related_work}
\paragraph{Image Matting.}
\label{subsec:Image_Matting}
The image matting task was introduced in \cite{porter1984compositing}.
Unlike the image segmentation task that predicts a binary alpha value, the image matting problem aims to predict a high-precision alpha value in a continuous range.
Therefore, the matting problem is harder to solve, and most existing works are addressed under some conditions.
The most common condition is assuming a human-annotated trimap is given.
The trimap is annotated into three different regions: definitely foreground region, definitely background region, and unknown region.
The trimap serves to not only reduce the difficulty of the matting problem but also allows some user control over the results.

The traditional sampling-based methods~\cite{chuang2001bayesian,wang2007optimized,gastal2010shared,he2011global,shahrian2013improving} determine alpha values in unknown regions by combining sampled foreground and background pixels.
Another traditional approach is the propagation-based ~\cite{jianpoisson,grady2005random,levin2007closed,levin2008spectral,lee2011nonlocal,chen2013knn,chen2013image}, which propagates foreground and background pixels into unknown regions based on affinity scores.
Recently, deep learning-based methods achieved great success in image matting~\cite{cho2016natural,xu2017deep,lu2019indices,hou2019context,li2020natural,forte2020f} by training networks with a trimap input.  

Some approaches try to find good alternatives to human-annotated trimap.
Portrait matting~\cite{shen2016deep,zhu2017fast,ke2020green} can extract an alpha matte without any external input, but they are only applicable for human subjects. 
Background matting~\cite{sengupta2020background,lin2021real} proposes to take complete background information instead of trimap input and predicts high-resolution alpha matte. 
However, this method is hard to extend to general video matting because it can work only with a near-static background.
Mask guided matting \cite{yu2021mask} proposes to replace the trimap with a coarse binary mask that is more accessible.
All image matting methods mentioned above can be extended to video matting by applying frame-by-frame, but the constraint must be met for each frame (\eg, trimap for every frame).

\paragraph{Video Matting.}
\label{subsec:Video_Matting}
Early video matting methods largely extended traditional image matting methods by either extending the propagation temporally~\cite{Apostoloff2004,Choi2012,Li2013,Shahrian2014} or by sampling in other frames~\cite{Lee2010,Shahrian2014}.  While there was some work that would generate trimaps automatically by deriving them from segmentations~\cite{gong2010,tang2010} or using interpolation~\cite{Chuang2002} or propagation~\cite{Lee2010}, these were computed independently from the video matting method. Bai \etal~\cite{Bai2011} propagates trimaps based on predicted alpha mattes. Tang \etal~\cite{tang2012} would use the alpha matte of one frame to help predict the trimap of the next frame, but did so by using the alpha matte to compute a binary segmentation from which a new trimap would be computed.

Recently, large-scale video matting datasets have been proposed~\cite{sun2021deep,zhang2021attention,lin2021real} and they fueled video matting researches based on deep learning. 
Along with a large benchmark dataset, Sun~\etal~\cite{sun2021deep} proposed a two-stage deep learning-based method by decoupling video matting into trimap generation and alpha matting.
They mainly focused on the alpha matting stage to achieve temporally coherent prediction by learning temporal feature alignment~\cite{wang2019edvr} and fusion~\cite{woo2018cbam}.
Concurrently, Zhang~\etal~\cite{zhang2021attention} also approach the problem with a similar motivation.
To be specific, \cite{zhang2021attention} proposed a temporal aggregation module based on the guided contextual attention block~\cite{li2020natural} to maintain temporal consistency during the alpha prediction. 
They also released a large-scale benchmark dataset, VideoMatting108~\cite{zhang2021attention}.

However, both latest learning-based methods~\cite{sun2021deep,zhang2021attention} overlooked the challenge in trimap propagation. 
Their temporally consistent results can be achieved only if the trimap is accurately generated for every frame. 
To prevent this issue, additional trimaps need to be manually annotated for several frames which makes the solutions less practical. 
Although there are trimap-free methods \cite{lin2022robust,sun2021modnet}, those work only on human video and cannot generate user-desired results.

Toward robust and practical trimap-based video matting, we revisit the traditional video matting studies~\cite{Bai2011,tang2012}, where propagating a trimap based on the predicted alpha matte greatly improves the temporal coherency.
Inspired by those studies while continuing the success of learning-based video matting, we propose OTVM that jointly learns trimap-based alpha prediction and alpha-based trimap propagation.
Our method shows that high-quality and temporally consistent alpha prediction in video is achievable using only a single trimap input.

\begin{figure}[t]
\centering
\includegraphics[width=1.\linewidth]{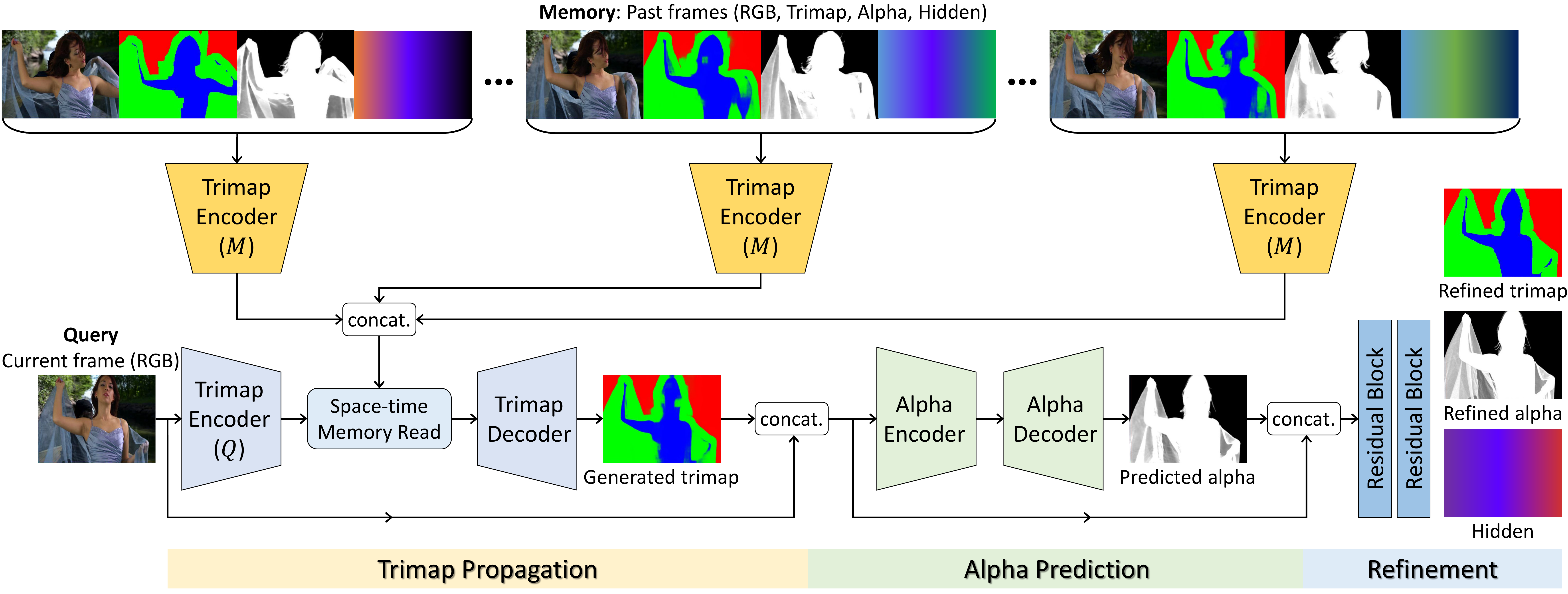}
\vspace{-0.5cm}
\caption{
\textbf{Overall architecture of OTVM}.
Our trimap propagation network is inspired by space-time memory networks~\cite{Oh_2019_ICCV}. 
The network predicts a trimap based on the information from the previous frames and predictions that are embedded by the trimap memory encoder.
From the given or generated trimap, we initially predict the alpha matte using our alpha prediction network inspired by~\cite{forte2020f}.
Then we refine the generated trimap and predicted alpha matte via two light-weighted residual blocks.
The refined trimap, alpha matte, and hidden features (dimension = 16) are fed to the trimap memory encoder so that they can be used for the next frames as a new memory.
The framework is trained in an end-to-end manner.
We further illustrate the details of each module in the supplementary material.
\vspace{-0.2cm}
}
\label{fig:overall_architecture}
\end{figure}

\secspaceA
\section{Method}
\secspaceB
\label{sec:method}

The overall architecture of OTVM is illustrated in \cref{fig:overall_architecture}.
Our method aims to perform video matting robustly with only a single trimap.
From the given user-provided trimap at the first frame, we sequentially predict a trimap and an alpha matte for every frame in a video sequence. 
Starting from the user-provided trimap, we first predict the initial alpha matte by feeding the RGB frame and trimap to our alpha prediction network. 
Then, a lightweight refinement module is followed to correct errors in alpha matte and trimap, resulting in a refined alpha matte and trimap.
The refinement module also produces a hidden latent feature map, and all the outputs from the refinement module are encoded as memory by the trimap memory encoder for trimap propagation.
On the next frame, our trimap propagation module predicts the trimap by reading relevant information from the memory. 
This procedure -- alpha prediction, refinement, and trimap propagation -- is repeated until the end of the video sequence.

Note that recent works~\cite{zhang2021attention,sun2021deep} also generate trimaps and predict alpha mattes, but our approach is completely different from those.
In the existing works, the trimap propagation and alpha matting are totally decoupled.
They tried to propagate trimap naively (\ie without consideration of the challenge introduced in \cref{fig:fig2}), resulting in inaccurate trimap propagation.
Therefore, multiple ground-truth (GT) trimaps should be provided to achieve good results.
In contrast, our OTVM can extract accurate alpha matte results even with a single human-annotated trimap, thanks to our joint modeling.

\subsecspaceA
\subsection{Alpha Prediction with Trimap}
\subsecspaceB
\label{subsec:Alpha_Matting_with_an_Unperfect_Trimap}

Given an RGB image and (either a propagated or user-provided) trimap, we predict an alpha matte. 
Here, we opt for the state-of-the-art image-based matting network, FBA \cite{forte2020f}, to simplify the problem. 
The alpha matting network is an encoder-decoder architecture. 
The alpha encoder first takes a concatenation of the RGB and trimap along the channel dimension as input.
Then, the resulting pyramidal features of the alpha encoder are fed into the alpha decoder that produces an alpha matte.
To exploit the advantage of the coupled network trained end-to-end, we directly use the soft trimap from the trimap propagation module without binarization when a propagated trimap is given.

We can use any alpha matting network, however, we empirically observe that advanced alpha networks (\eg video alpha networks) make our framework complex and hinder end-to-end training given limited training data for video matting, while the latest image-based alpha networks work surprisingly well as long as a reliable trimap and end-to-end training are provided. Therefore, we take the simple image-based alpha prediction model and focus on developing the joint framework that can reliably propagate the trimaps.

\subsecspaceA
\subsection{Alpha-Trimap Refinement}
\subsecspaceB
In our video matting setting, which takes only one GT trimap, naively propagating the trimap may result in severe drifting as depicted in Fig.~\ref{fig:fig2}. 
To take the advantage of our coupled framework, we have an additional refinement module following the alpha network to provide refined information to the trimap propagation module afterward.
The refinement module is light-weighted as it is composed of two residual blocks.
The refinement module takes all available information for the current frame: an input RGB frame, the generated trimap, the predicted alpha matte, and the alpha decoder's latent features.
Then, the module produces an updated trimap and a refined alpha matte along with unconstrained hidden features.
The hidden features are intended for information that cannot be expressed in the form of trimap and alpha.
These features are learnable through end-to-end training. 
All the outputs of the refinement module will be used for trimap propagation.

\subsecspaceA
\subsection{Trimap Propagation with Alpha}
\subsecspaceB
\label{subsec:Trimap_Propagation}
To propagate trimaps, we repurpose a state-of-the-art video segmentation network, space-time memory network (STM)~\cite{Oh_2019_ICCV}, with important modifications.
In the original STM~\cite{Oh_2019_ICCV}, input images and corresponding masks in the past frames are set to the memory, while the image at the current frame is set to the query.
Then, the memory and query are embedded through two independent ResNet50~\cite{b23} encoders.
The embedded memory and query features are fed to the space-time memory read module.
In the module, dense matching is performed and then a value of memory is retrieved based on the matching similarity.
The decoder takes the retrieved memory value and query feature and then outputs an object mask.
This approach can effectively exploit rich features of the intermediate frames and achieve state-of-the-art performance in video binary segmentation.

STM~\cite{Oh_2019_ICCV} simply can be extended from binary mask to trimap by increasing the input channel dimension of the memory encoder and the output channel dimension of the decoder.
However, there is a fundamental limitation to apply STM directly for trimap propagation.
In the binary mask, the foreground region and background region can be estimated by propagating from past binary masks.
The trimap, however, cannot be estimated only with propagation because the unknown regions are frequently changed by the view of the foreground object (see \cref{fig:fig2}) and trimap-only supervision does not provide a consistent clue for the changes.
To address this problem, we additionally impose rich cues for generating the trimap into the memory encoder.
Since the trimap has been determined by the alpha matte, it effectively helps to learn for trimap generation.
We additionally impose a hidden feature extracted from the refinement module.
With the hidden features, any errors can be easily propagated backward at training time, resulting in stable training.
By imposing those into the memory encoder, we significantly reduce errors that occurred by drifting of the unknown regions.

\subsecspaceA
\subsection{End-to-End Training}
\subsecspaceB
\label{subsec:End-to-End_Training}
To make OTVM work, it is critical to train the model end-to-end because each module depends on each other's outputs. 
However, video matting data are extremely difficult to annotate and existing video supervisions are not sufficient to train the model directly.
As a practical solution, we train each module stage-wise then fine-tune the whole network in an end-to-end manner.
First, we initialize the trimap propagation and the alpha matting modules with the pretrained weights of off-the-shelf STM~\cite{Oh_2019_ICCV} and FBA~\cite{forte2020f}, respectively.
Specifically, both pretrained models leverage ImageNet~\cite{b19}. 
In addition, the STM is trained using image segmentation datasets~\cite{b42,b43,hariharan2011semantic,shi2015hierarchical,cheng2014global} and video object segmentation datasets~\cite{pont20172017,xu2018youtube}. 
The FBA is trained on the Adobe Image Matting (AIM) dataset~\cite{xu2017deep}.
After the initialization, we pretrain our network modules in three stages (from Stage 1 to Stage 3) on the AIM dataset~\cite{xu2017deep} and then finetune the whole model end-to-end on either VideoMatting108~\cite{zhang2021attention} or DVM~\cite{sun2021deep}, depending on the target evaluation benchmark (Stage 4).

\paragraph{Stage 1: Training the alpha matting module and trimap propagation module separately.}~
As two modules depend on each other, if we train the alpha matting module and the trimap propagation module simultaneously from scratch, this can lead the model to either poor convergence or simply memorizing training data (\ie, overfitting). 
It is because both modules cannot learn meaningful features from almost randomly initialized input data which is as the output of other modules.
Therefore, we first separately train two modules without the connections between two. 
Specifically, we train the alpha matting module with GT trimaps and train the trimap propagation module without taking inputs of an alpha matte and hidden features.

\paragraph{Stage 2: Training the alpha matting and refinement modules with propagated trimaps.}~We train the alpha matting model and refinement modules together while the trimap propagation module is frozen.
This stage enables the refinement module to take a soft and noisy trimap as input and learn to predict accurate trimap and alpha matte.

\paragraph{Stage 3: Training the trimap propagation module.}~In the trimap propagation module, we activate all input layers for alpha matte and hidden features. 
Then, we train the trimap propagation module while the parameters for the remainders -- alpha prediction and refinement -- are frozen.
In this stage, we leverage not only the loss from the predicted trimap but also the losses from the alpha prediction.
This enables the trimap propagation module to predict a more reliable trimap for estimating the alpha matte.
While we are not updating the alpha network and refinement module in this stage, we can leverage the gradients from their losses for updating the trimap propagation module.

\paragraph{Stage 4: End-to-end training.}~Finally, we train the whole network end-to-end using a video matting dataset.
With the stage-wise pretraining, we can effectively leverage both image and video data, and achieve stable performance improvement at the end-to-end training.

\subsecspaceA
\subsection{Training Details}
\subsecspaceB
\label{subsec:Training_Details}
\paragraph{Data preparation.}~During training, we randomly sampled three temporally ordered foreground and background frames from each video sequence.
If an image dataset (\eg, AIM~\cite{xu2017deep}) is used, we simulate three video frames from a pair of foreground and background images by applying three different random affine transforms into both foreground and background images.
The random affine transforms include horizontal flipping, rotation, shearing, zooming, and translation.
For each foreground and background frame, we randomly crop patches into $320 \times 320$, $480 \times 480$, or $640 \times 640$, centered on pixels in the unknown regions.
And then we resize the cropped patches into $320 \times 320$.
Additionally, we employ several augmentation strategies on both foreground and background frames: histogram matching between foreground and background colors, motion blur, Gaussian noise, and JPEG compression.
Then we composite foreground and background on-the-fly to generate an input frame.
The GT trimaps are generated by dilating the GT alpha matte with a random kernel size from $1 \times 1$ to $26 \times 26$.

\paragraph{Loss functions.}~We set objective functions for all outputs of the models, except for the hidden features.
For both initially predicted and refined trimaps, we use the cross-entropy loss to compare with the GT.
For the first frame where the GT trimap is provided as the input, we only apply the loss to the refined trimap. 
Ideally, there should be no change after refinement. 
We find penalizing any change after the refinement is helpful to prevent it from corrupting already accurate trimap.
For the alpha predictions, we leverage the temporal coherence loss~\cite{sun2021deep} and image matting losses used in FBA~\cite{forte2020f}.
Different from some previous methods that only compute alpha losses on unknown regions (\eg~\cite{xu2017deep}), we compute our losses on every pixel. 

In addition to trimap and alpha losses, we also employ losses for the foreground and background color predictions.  
We estimate foreground and background colors from the alpha decoder and refinement module following~\cite{forte2020f}.
We minimize all foreground and background losses used in~\cite{forte2020f}, and additionally employ temporal coherence loss on both foreground and background.
For the foreground color, we compute the losses only where an alpha value is greater than 0 because the exact foreground color is available only in those regions.
More detailed explanations of loss functions are given in the supplementary material.

\paragraph{Other training details.}
We opt for RAdam optimizer~\cite{liu2019radam} with a learning rate of 1e-5.
We drop the learning rate to 1e-6 once at 90\% iteration for each training stage.
We freeze all the batch normalization layers in the networks.
We used a mini-batch size of 4 and trained with four NVIDIA GeForce 1080Ti GPUs.
At the first pretraining stage, we trained the alpha matting model about 100,000 iterations and we trained the trimap propagation model about 400,000 iterations.
We trained about 50,000 iterations at each of the second and third training stages.
Finally, we trained about 80,000 iterations at the last end-to-end training stage.

\subsecspaceA
\subsection{Inference Details}
\subsecspaceB
\label{subsec:Inference_Details}
We used full-resolution inputs to achieve high-quality alpha matte results.
For the memory management in the trimap propagation module, we generally follow STM~\cite{Oh_2019_ICCV} that stores the first and the previous frames to the memory by default, and additionally saves new memory periodically. 
We add the intermediate frames to the memory for every 10 frames.
To avoid GPU memory overflow, we store only the last three intermediate frames and discard old frames.

\secspaceA
\section{Experiments}
\label{sec:Experiments}
\subsecspaceA
\subsection{Evaluation Datasets and Metrics}
\label{subsec:Evaluation_metrics}
\subsecspaceB
We present experimental results and analysis on two latest benchmarks, VideoMatting108~\cite{zhang2021attention} and DVM~\cite{sun2021deep}.

\paragraph{VideoMatting108 dataset~\cite{zhang2021attention}}
 includes 28 foreground video sequences paired with background video sequences in the validation set.
The evaluation is conducted with three different trimap settings: narrow, medium, and wide. 
The groundtruth trimaps are generated by discretizing the groundtruth alpha mattes into the trimaps, followed by dilating the unknown regions with different kernel sizes; $11 \times 11$ for narrow, $25 \times 25$ for medium, and $41 \times 41$ for wide.
This benchmark contains long video sequences and the average length of the videos is about 850 frames.
Therefore, predicting alpha mattes with only a single trimap at the first frame is challenging in this dataset.

\paragraph{DVM dataset~\cite{sun2021deep}} provides 12 foreground videos and 4 background videos for validation.
The validation set is constructed with an additional 50 foreground images from the AIM dataset~\cite{xu2017deep}.
In total, 62 foregrounds are composited with every 4 background videos, resulting in 248 test videos.
Following the DVM paper~\cite{sun2021deep}, we discard 106 foreground images from the AIM dataset~\cite{xu2017deep} and use the remaining 325 foreground images during our training because there are overlaps in the evaluation. 

\paragraph{Evaluation metrics.}
For a fair comparison, we follow the evaluation metrics from two large-scale video matting benchmarks~\cite{zhang2021attention,sun2021deep}.
To evaluate on VideoMatting108~\cite{zhang2021attention}, we compute SSDA (average sum of squared difference), MSE (mean squared error), MAD (mean absolute difference), dtSSD (mean squared difference of direct temporal gradients), and MESSDdt (mean squared difference between the warped temporal gradient)~\cite{erofeev2015perceptually}.
To evaluate on DVM~\cite{sun2021deep}, we compute SAD (sum of absolute difference), MSE, Grad (gradient error), Conn (connectivity error), dtSSD, and MESSDdt. For all computed metrics, lower is better.

In the original works~\cite{zhang2021attention,sun2021deep}, the metrics are computed only on the unknown regions of the GT trimaps.
We follow this rule for fair comparisons with the existing methods.
However, using only the unknown region cannot capture the errors in the foreground and background regions that occurred by inaccurate trimap propagation, which is important for evaluating the performance of end-to-end video matting methods.
Therefore, we present \textit{the modified versions of the metrics} that compute the scores on the full-frames, suffixed with ``-V''. The modified metrics are used for our analysis and ablation experiments.

\begin{table}[t]
\caption{Analysis experiments on VideoMatting108 validation set. For all experiments, we use 1-trimap setting where GT trimap is given only at the first frame. ``-V'' denotes the error has been computed in all regions of the frames (see \cref{subsec:Evaluation_metrics}).}
\label{tab1}
\setlength{\tabcolsep}{6pt}
\centering
\vspace{-3mm}\mytabular{(a) \textbf{Joint modeling}.}\\
\mytabular{
\setlength{\tabcolsep}{7.1pt}
\begin{tabular}{cc|ccccc}
\toprule
Model                    & Training method & SSDA-V          & MSE-V          & MAD-V           & dtSSD-V         & MESSDdt-V         \\
\midrule
\multirow{2}{*}{STM+FBA} & decoupled       & 83.61          & 10.62         & 22.12          & 36.31          & 3.45          \\
                         & joint      & \textbf{75.36} & \textbf{9.40} & \textbf{21.01} & \textbf{29.64} & \textbf{2.74}             \\
                         \bottomrule
\end{tabular}
}
\\ \vspace{2mm}\mytabular{(b) \textbf{Stage-wise training}.}\\
\mytabular{
\setlength{\tabcolsep}{8.1pt}
\begin{tabular}{cc|ccccc}
\toprule
Model                    & Training method & SSDA-V          & MSE-V          & MAD-V           & dtSSD-V         & MESSDdt-V   \\
\midrule
\multirow{2}{*}{OTVM}    & joint      & 72.07          & 9.29          & 20.81          & 30.20          & 2.79             \\
                         & joint + stage-wise      & \textbf{54.67} & \textbf{2.61} & \textbf{13.02} & \textbf{29.87} & \textbf{1.78}     \\
                         \bottomrule
\end{tabular}
}
\\ \vspace{2mm}\mytabular{(c) \textbf{Ablation on training stages}. Each training stage is described in \cref{subsec:End-to-End_Training}.}\\
\mytabular{
\setlength{\tabcolsep}{5.3pt}
\begin{tabular}{cccc|ccccc}
\toprule
\multicolumn{4}{c|}{Train stages}            & \multirow{2}{*}{SSDA-V} & \multirow{2}{*}{MSE-V} & \multirow{2}{*}{MAD-V} & \multirow{2}{*}{dtSSD-V} & \multirow{2}{*}{MESSDdt-V}  \\
Stage 1    & Stage 2    & Stage 3    & Stage 4         &                        &                       &                       &                         &                         \\
\midrule
           &            &            & \checkmark & 87.31                  & 11.16                 & 23.35                 & 33.29                   & 3.15                        \\
\checkmark &            &            & \checkmark & 76.55                  & 9.68                  & 23.10                 & 31.64                   & 3.14                        \\
\checkmark & \checkmark &            & \checkmark & 75.33                  & 9.54                  & 22.30                 & 31.44                   & 3.06                        \\
\checkmark & \checkmark & \checkmark & \checkmark & \textbf{54.67}         & \textbf{2.61}         & \textbf{13.02}        & \textbf{29.87}          & \textbf{1.78}             \\
\bottomrule
\end{tabular}
}
\\ \vspace{2mm}\mytabular{(d) \textbf{Ablation on modules}. The modules are depicted in Fig.~\ref{fig:overall_architecture}.} \\
\mytabular{
\setlength{\tabcolsep}{2.5pt}
\begin{tabular}{cccc|cccccc}
\toprule
\multicolumn{2}{c}{\begin{tabular}[c]{@{}c@{}}Refinement module\\ (output)\end{tabular}} & \multicolumn{2}{c|}{\begin{tabular}[c]{@{}c@{}}Trimap module\\ (input)\end{tabular}} & \multirow{2}{*}{SSDA-V} & \multirow{2}{*}{MSE-V} & \multirow{2}{*}{MAD-V} & \multirow{2}{*}{dtSSD-V} & \multirow{2}{*}{MESSDdt-V} & \multirow{2}{*}{{\begin{tabular}[c]{@{}c@{}}Time\\(sec/frame)\end{tabular}}} \\
Alpha                                  & Trimap                                & Alpha                                & Hidden                               &                       &                      &                      &                        & &                        \\
\midrule
                                       &                                       &                                      &                                      & 83.56                  & 12.17                 & 23.59                 & 30.11                   & 2.80     & 0.799                   \\
\checkmark                             &                                       &                                      &                                      & 83.51                  & 11.26                 & 22.49                 & 31.78                   & 2.89     & 0.951                   \\
                                       & \checkmark                            &                                      &                                      & 78.70                  & 9.80                  & 21.51                 & 29.97                   & 2.80     & 0.946                   \\
\checkmark                             & \checkmark                            &                                      &                                      & 77.58                  & 10.55                 & 22.42                 & 31.04                   & 2.98      & 0.952                  \\
\checkmark                             & \checkmark                            & \checkmark                           &                                      & 62.47                  & 4.27                  & 15.06                 & 29.98                   & 1.85     & 0.955                   \\
\checkmark                             & \checkmark                            & \checkmark                           & \checkmark                           & \textbf{54.67}         & \textbf{2.61}         & \textbf{13.02}        & \textbf{29.87}          & \textbf{1.78}      & 0.964        \\
\bottomrule
\end{tabular}
}
\\\vspace{2mm}\mytabular{(e) \textbf{Different image matting backbones}.} \\
\mytabular{
\setlength{\tabcolsep}{9.1pt}
\begin{tabular}{cl|ccccc}
\toprule
Backbone                                  & Model               & SSDA-V          & MSE-V           & MAD-V           & dtSSD-V         & MESSDdt-V         \\
\midrule
\multirow{2}{*}{DIM~\cite{xu2017deep}}    & STM+DIM  & 102.77         & 12.58          & 28.35          & 47.13          & 5.14             \\
                                          & OTVM                 &  \textbf{92.30}          & \textbf{11.04} & \textbf{25.98} & \textbf{39.27} & \textbf{4.21}      \\
                                          \midrule
\multirow{2}{*}{GCA~\cite{li2020natural}} & STM+GCA  & 100.22         & 12.25          & 27.33          & 41.72          & 4.36             \\
                                          & OTVM                 & \textbf{84.73} & \textbf{10.04} & \textbf{24.23} & \textbf{36.02} & \textbf{3.46}     \\
                                          \bottomrule
\end{tabular}
}
\\\vspace{2mm}\mytabular{
\begin{tabular}{c}
(f) \textbf{Trimap propagation}.
``-T'' is presented to estimate trimap quality and denotes that\\ the unknown region in GT trimap has been modified (see \cref{subsec:Analysis_Experiments}).
\end{tabular}
}
\mytabular{
\setlength{\tabcolsep}{10pt}
\begin{tabular}{l|ccc}
\toprule
Method                                                & \begin{tabular}[c]{@{}c@{}}Precision-T \end{tabular} & \begin{tabular}[c]{@{}c@{}}Recall-T \end{tabular} & Average \\
\midrule
Decoupled   STM~\cite{sun2021deep,zhang2021attention} & 96.98                                                                       & 93.58                                                                    & 95.28   \\
OTVM                                                  & \textbf{98.17} & \textbf{95.92} & \textbf{97.05}  \\
\bottomrule
\end{tabular}
}
\vspace{-5mm}
\end{table}

\subsecspaceA
\subsection{Analysis Experiments}
\label{subsec:Analysis_Experiments}
\subsecspaceB
To validate our hypotheses experimentally, we conduct a set of analysis experiments. We use the one-trimap setting, where GT trimap is given only at the first frame. All these analysis experiments are conducted on the VideoMatting108 benchmark with the medium trimap setting.

\paragraph{Effectiveness of the joint modeling.}~We first validate the importance of the joint modeling of trimap propagation and alpha prediction in video matting.
For this purpose, we design a simple baseline model for video matting, STM+FBA, that cascades STM~\cite{Oh_2019_ICCV} for trimap propagation and FBA~\cite{forte2020f} for alpha prediction from the propagated trimap.
Note that we do not use OTVM because our proposals (\ie, trimap and alpha refinement and using hidden features) are only applicable for the joint modeling.
To show the effect of joint modeling, we train the baseline model with two different training strategies. 
One is obtained by training trimap propagation (STM) and alpha prediction (FBA) separately (\ie, decoupled), and the other is by training both modules jointly (\ie, joint).
As shown in \cref{tab1}(a), the joint modeling greatly improves the alpha matte quality in the practical one-trimap scenario.

\paragraph{Efficacy of the stage-wise training.}~We evaluate the importance of the proposed stage-wise training in OTVM and summarize the result in \cref{tab1}(b).
When we activate all input layers for alpha matte and hidden features and end-to-end train OTVM from the pretraining stage (\ie, joint),
the model marginally surpasses the STM+FBA (joint). If our stage-wise training is applied, OTVM significantly outperforms STM+FBA (joint). The results demonstrate the superiority of our stage-wise training.

\paragraph{Efficacy of each training stage.}~In this experiment, we use OTVM and validate our training strategy. The result is summarized in \cref{tab1}(c).
To learn the trimap propagation model with the hidden features of the refinement module, we applied the last training stage (\ie, Stage 4: end-to-end training) for all cases.
As shown in the table, each pretraining stage consistently improves performance.
The results demonstrate that our training strategy, stage-wise pretraining then end-to-end finetuning, is effective.

\paragraph{Effectiveness of the proposed modules.}~We thoroughly evaluate our proposed modules in \cref{tab1}(d).
As shown in the table, the alpha matte and trimap refinements lead to complementary performance improvement.
More importantly, feeding the alpha matte and hidden features of the refinement module to the trimap propagation module significantly improves performance with small extra computation.
The result demonstrates the importance of reliable trimap propagation in the practical video matting setting. It also shows that our joint learning framework and the proposed modifications effectively address the problem.

\paragraph{Applicability to different image matting networks.}~\cref{tab1}(e) compares the results with other image matting backbones. We tested DIM~\cite{xu2017deep} and GCA~\cite{li2020natural}. As shown in the table, OTVM consistently leads to performance improvements over the corresponding baselines.

\paragraph{Accuracy of trimap propagation.}~We study the effect of OTVM on trimap propagation. In \cref{tab1}(f), we compare OTVM with recent video matting methods~\cite{sun2021deep,zhang2021attention} that naively propagate trimap using a single STM~\cite{Oh_2019_ICCV}.
Ideally, the unknown area in a trimap needs to cover the entire soft matte area in the GT matte while being tight enough not to be trivial.
To measure the quality of trimaps, we present two metrics: (1) \textit{Precision-T}, precision of the estimated unknown area compared with the widely dilated GT unknown (dilation kernel size of $41 \times 41$) and (2) \textit{Recall-T}, recall of the estimated unknown area compared with the minimum GT unknown (\ie, no dilation of the GT unknown regions). As shown in \cref{tab1}(f), OTVM significantly outperforms the state-of-the-art approach and achieves high precision and high recall.

\paragraph{More analysis in the supplementary material.}~We present additional results on the input of the trimap encoder, a visual analysis of the hidden feature, the effect of the refinement module, an analysis of runtime and GPU memory consumption, an analysis of temporal stability, and quantitative results with image matting metrics (\ie without ``-V'') in the supplementary material.

\begin{table}[t]
\setlength{\tabcolsep}{7pt}
\caption{Comparison with state-of-the-art methods on public benchmarks. The trimap setting indicates how many GT trimaps are given as input, \ie, ``full-trimap" for all frames, ``20/40-frame" for every 20/40th frames, ``1-trimap" for only at the first frame.}
\label{table:comparison}
\centering
\vspace{-0.2cm} \mytabular{
\begin{tabular}{c}
(a) \textbf{Comparison on VideoMatting108 validation set}.
Results for other methods are directly copied from~\cite{zhang2021attention}.\\
$^\dagger$ denotes our reproduced results using our training setup.
\end{tabular}
} \\
\mytabular{
\begin{tabular}{cl|ccccc}
\toprule
Trimap   Setting             & Methods                                          & SSDA           & MSE           & MAD            & dtSSD          & MESSDdt         \\
\midrule
\multirow{5}{*}{full-trimap} & DIM~\cite{xu2017deep}                            & 61.85          & 9.99          & 44.38          & 34.55          & 2.82          \\
                             & IndexNet~\cite{lu2019indices}                    & 58.53          & 9.37          & 43.53          & 33.03          & 2.33          \\
                             & GCA~\cite{li2020natural}                         & 55.82          & 8.20          & 40.85          & 31.64          & 2.15          \\
                             & TCVOM (GCA)~\cite{zhang2021attention}                   & 50.41          & 7.07          & 37.65          & 27.28          & 1.48          \\
                             & TCVOM   (FBA)$^\dagger$~\cite{zhang2021attention}       & \textbf{39.76} & \textbf{4.01} & \textbf{28.68} & \textbf{22.93} & \textbf{1.06} \\
                             \midrule
\multirow{4}{*}{1-trimap}    & STM + TCVOM (GCA)~\cite{zhang2021attention}             & 77.23          & 22.15         & 57.40          & 32.18          & 2.97          \\
                             & STM + TCVOM   (FBA)$^\dagger$~\cite{zhang2021attention} & 69.96          & 19.80         & 51.21          & 29.76          & 2.72          \\
                             & STM +   FBA$^\dagger$~\cite{forte2020f}          & 70.63          & 20.18         & 51.20          & 31.00          & 2.86          \\
                             & OTVM                                             & \textbf{50.51} & \textbf{8.58} & \textbf{37.16} & \textbf{28.28} & \textbf{1.63} \\
                             \bottomrule
\end{tabular}
}
\\ \vspace{2mm} \mytabular{(b) \textbf{Comparison on DVM validation set}.
Results for other methods are directly copied from~\cite{sun2021deep}.}\\
\mytabular{
\begin{tabular}{cl|cccccc}
\toprule
Trimap Setting & Methods & SAD          & MSE          & Grad           & Conn         & dtSSD & MESSDdt       \\
\midrule
\multirow{5}{*}{full-trimap} & DIM~\cite{xu2017deep}               & 54.55          & 0.030          & 35.38          & 55.16          & 23.48          & 0.53          \\
                                                                        & IndexNet~\cite{lu2019indices}       & 53.68          & 0.028          & 27.52          & 54.44          & 19.50          & 0.49          \\
                                                                        & Context-Aware~\cite{hou2019context} & 51.78          & 0.027          & 28.57          & 49.46          & 19.37          & 0.50          \\
                                                                        & GCA~\cite{li2020natural}            & 47.49          & 0.022          & 26.37          & 45.23          & 18.36          & 0.33          \\
                                                                        & DVM~\cite{sun2021deep}              & \textbf{40.91} & \textbf{0.014} & \textbf{19.02} & \textbf{40.58} & \textbf{15.11} & \textbf{0.25} \\
\midrule
\multirow{2}{*}{20-frame}                                               & DVM~\cite{sun2021deep}              & 43.66          & 0.016          & 26.39          & 42.23          & 16.34          & 0.28          \\
                                                                        & OTVM                                & \textbf{37.90} & \textbf{0.013} & \textbf{19.13} & \textbf{36.48} & \textbf{14.76} & \textbf{0.22} \\
\midrule
\multirow{2}{*}{40-frame}                                               & DVM~\cite{sun2021deep}              & 52.85          & 0.026          & -              & -              & 19.23          & -            \\
                                                                        & OTVM                                & \textbf{38.24} & \textbf{0.014} & \textbf{19.29} & \textbf{36.83} & \textbf{14.82} & \textbf{0.22} \\
\midrule
\multirow{2}{*}{1-trimap}                                               & DVM~\cite{sun2021deep}              & 65.33          & 0.039          & -              & -              & 35.46          & -            \\
                                                                        & OTVM                                & \textbf{41.02} & \textbf{0.017} & \textbf{20.17} & \textbf{39.79} & \textbf{14.85} & \textbf{0.25} \\
                                                                        \bottomrule
\end{tabular}
}
\vspace{-5mm}
\end{table}

\subsecspaceA
\subsection{Comparison with State-of-the-Art Methods}
\label{subsec:Comparison_on_Benchmark_Sets}
\subsecspaceB
\subsubsection{VideoMatting108~\cite{zhang2021attention}.}~\cref{table:comparison}(a) shows quantitative evaluation results on the VideoMatting108 validation set with medium trimap setting. Results with other trimap settings (\ie, narrow and wide) are provided in the supplementary material.
In the experiment, we measured errors only on the unknown regions for a fair comparison with \cite{zhang2021attention}.
We report two results depending on whether GT trimap is used for all frames or for the first frame.
For testing other methods on 1-trimap setting, we used the baseline trimap propagation model we trained to generate trimaps.
For a fair comparison with TCVOM~\cite{zhang2021attention}, which uses GCA~\cite{li2020natural} for a backbone network, we additionally show FBA backbone results.
When a single trimap is used, OTVM significantly outperforms all other methods, demonstrating the superiority of our approach.

\paragraph{DVM~\cite{sun2021deep}.}~We conduct quantitative evaluation on the DVM benchmark and \cref{table:comparison}(b) shows the results.
We computed errors only on the unknown regions according to the official metric.
\cref{table:comparison}(b) shows that OTVM largely surpasses DVM in 1-trimap setting.
Furthermore, OTVM using only 1-trimap (SAD 41.02) achieves comparable performance with DVM using full-trimap (SAD 40.91).

\begin{figure}[t]
\centering
\includegraphics[width=.8\linewidth]{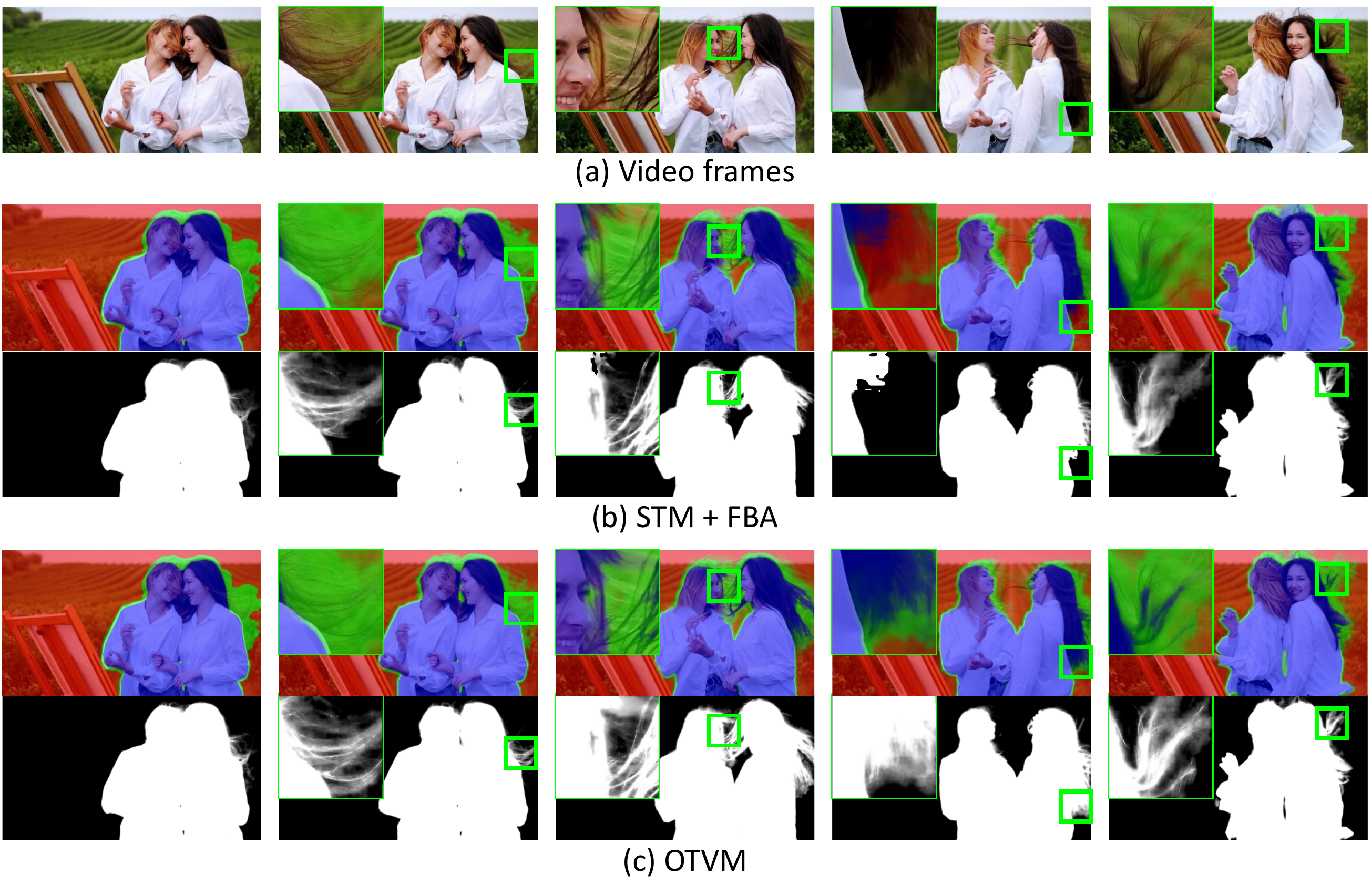}
\vspace{-0.5cm}
\caption{
Qualitative comparison on a real-world video sequence.
As shown in zoomed trimap, OTVM generates more accurate and fine trimaps, \eg, sharply separates hairs in the foreground region, but STM roughly predicts the regions as unknown.
Therefore, we can extract high-quality alpha matte results.
\vspace{-0.3cm}
}
\label{fig:qualitative_results_real_video}
\end{figure}

\paragraph{Qualitative results on real-world videos.}
\label{subsec:Comparison_on_Real_Videos}~\cref{fig:qualitative_results_real_video} shows qualitative results on a real-world video. We compare OTVM with the cascaded STM and FBA model, denoted by STM+FBA.
As shown in the figure, the STM+FBA model cannot sharply separate foreground and unknown regions on the object boundary. The cascade baseline model fails to predict accurate trimaps in the challenging scenes, resulting in poor alpha mattes.
In contrast, OTVM predicts trimaps reliably, resulting in accurate alpha mattes.
More qualitative results are provided in the supplementary material.
In addition, we provide full-frame results online: \url{https://youtu.be/qkda4fHSyQE}.

\begin{figure}[t]
\centering
\includegraphics[width=.8\linewidth]{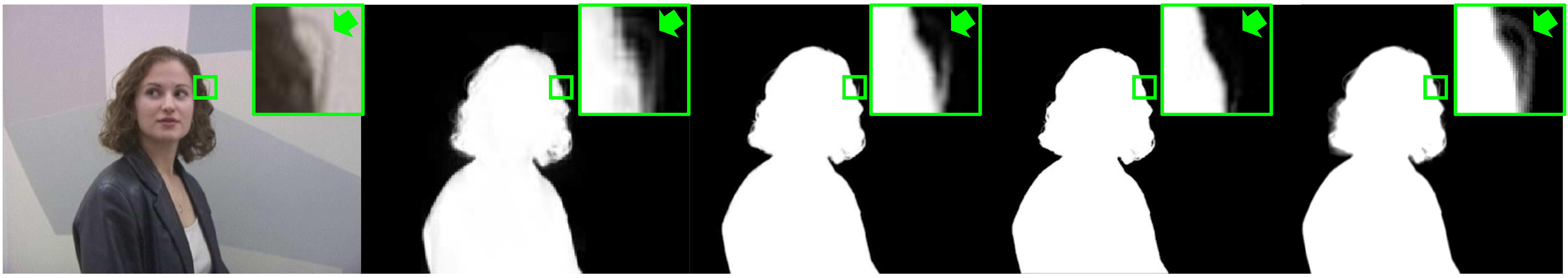}\\
\vspace{-0.2cm}
\raggedright{\scriptsize \hspace{35pt} \textit{Amira}, frame 43 \hspace{11pt} SVM~\cite{eisemann2009spectral} \hspace{8pt} Nonlocal~\cite{Choi2012} \hspace{0pt} \textls[-90]{KNN Vid. Mat.~\cite{Li2013}} \hspace{0pt} OTVM}
\vspace{-0.3cm}
\caption{
Comparison with non-deep learning methods.
\vspace{-0.5cm}
}
\label{fig:qualitative_results_amira}
\end{figure}

\paragraph{Comparison with non-deep learning methods.}~\cref{fig:qualitative_results_amira} shows a comparison with \cite{eisemann2009spectral,Choi2012,Li2013} on the \textit{Amira} benchmark~\cite{Chuang2002}.
The benchmark~\cite{Chuang2002} does not provide the GT alpha matte and we took the results of the previous methods from \cite{Li2013,eisemann2009spectral}.
As shown in the figure, SVM~\cite{eisemann2009spectral} result is noisy. Both Nonlocal~\cite{Choi2012} and KNN Video Matting~\cite{Li2013} fail to predict hair strand details.
In contrast, OTVM predicts the precise alpha matte.

\subsecspaceA
\subsection{Limitations}
\begin{figure}[t]
\centering
\vspace{-0.2cm}
\includegraphics[width=\linewidth]{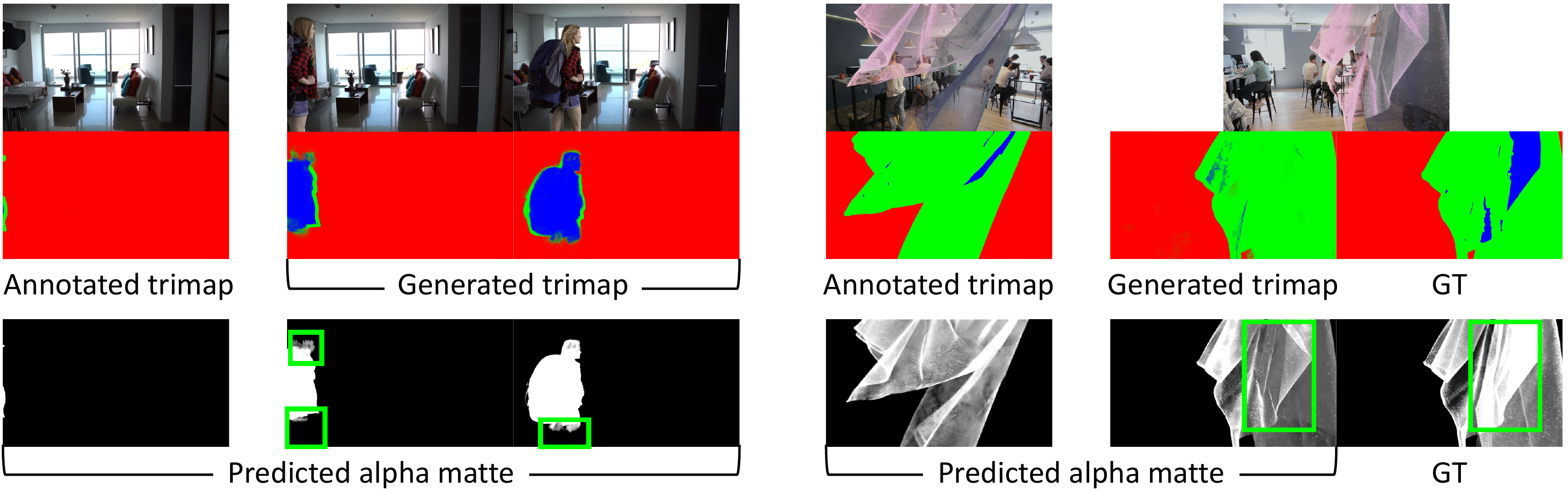}\\
\vspace{-0.15cm}
\raggedright{\scriptsize \hspace{60pt} (a) Case1 \hspace{143pt} (b) Case2}
\vspace{-0.3cm}
\caption{
Limitations. We indicate error areas with green boxes. (a) The user-annotated trimap contains almost no object information. 
(b) There is no strong signal for the foreground in the given trimap.
\vspace{-0.5cm}
}
\label{fig:limitations}
\end{figure}

\subsecspaceB
\label{subsec:Limitations}
Since our framework takes only a single user-annotated trimap, not only the input trimap quality but also the rich cues of the object in the frame are important.
In \cref{fig:limitations}(a), OTVM struggles to generate accurate trimaps if the user-annotated frame contains almost no object information, resulting in failure to predict the alpha matte.
In \cref{fig:limitations}(b), although the object is presented in the annotated frame, we may struggle to predict precise alpha mattes if there is no strong signal for the foreground object in the given trimap.

\subsecspaceA
\section{Conclusion}
\secspaceB
\label{sec:Conclusion}
In this paper, we present a new video matting framework that only needs a single user-annotated trimap.
In contrast to the recent decoupled methods that focus on alpha prediction given the trimaps, we propose a coupled framework, OTVM, that performs trimap propagation and alpha prediction jointly. OTVM with one user-annotated trimap significantly outperforms the previous works in the same setting and even achieves comparable performance with the previous works using full-trimaps as input.
OTVM is simple yet effective and works robustly in the practical one-trimap scenario. We hope that our research motivates follow-up studies and leads to practical video matting solutions.

\paragraph{Acknowledgements.}
This research was supported in part by the Yonsei Signature Research Cluster Program of 2022 (2022-22-0002). This research was also supported in part by the KIST Institutional Program (Project No. 2E31051-21-204).

\clearpage
%
%
\bibliographystyle{splncs04}
\bibliography{egbib}

\clearpage
\crefname{section}{Sec.}{Secs.}
\Crefname{section}{Section}{Sections}
\Crefname{table}{Table}{Tables}
\crefname{table}{Table}{Tables}
\crefname{appendix}{Sec.}{Secs.}
\Crefname{appendix}{Section}{Sections}

\makeatletter
\DeclareRobustCommand\onedot{\futurelet\@let@token\@onedot}
\def\@onedot{\ifx\@let@token.\else.\null\fi\xspace}

\def\eg{\emph{e.g}\onedot} \def\Eg{\emph{E.g}\onedot}
\def\ie{\emph{i.e}\onedot} \def\Ie{\emph{I.e}\onedot}
\def\cf{\emph{cf}\onedot} \def\Cf{\emph{Cf}\onedot}
\def\etc{\emph{etc}\onedot} \def\vs{\emph{vs}\onedot}
\def\wrt{w.r.t\onedot} \def\dof{d.o.f\onedot}
\def\iid{i.i.d\onedot} \def\wolog{w.l.o.g\onedot}
\def\etal{\emph{et al}\onedot}
\makeatother

\newcommand{\beginsupplement}{
	\setcounter{table}{0}
	\renewcommand{\thetable}{S\arabic{table}}%
	\setcounter{figure}{0}
	\renewcommand{\thefigure}{S\arabic{figure}}%
	\setcounter{equation}{0}
	\renewcommand{\theequation}{S\arabic{equation}}
}

\title{Supplementary Material: \\ One-Trimap Video Matting} 

\titlerunning{One-Trimap Video Matting}
%
\author{Hongje Seong\inst{1,}\thanks{This work was done during an internship at Adobe Research.} \and Seoung Wug Oh\inst{2} \and Brian Price\inst{2} \and \\ Euntai Kim\inst{1} \and Joon-Young Lee\inst{2}}
\authorrunning{H. Seong, S. W. Oh, B. Price, E. Kim, and J.-Y. Lee}
%
\institute{Yonsei University, Seoul, Korea. \email{\{hjseong,etkim\}@yonsei.ac.kr} \and Adobe Research, San Jose, CA, USA. \email{\{seoh,bprice,jolee\}@adobe.com}}
\maketitle

\appendix
\beginsupplement

\section{Network Structure Details}
\label{sec:network_structure_details}
In this section, we describe detailed network structures for trimap propagation, alpha prediction, and alpha-trimap refinement.

\begin{figure}
\centering
\vspace{-5mm}
\includegraphics[width=1.\linewidth]{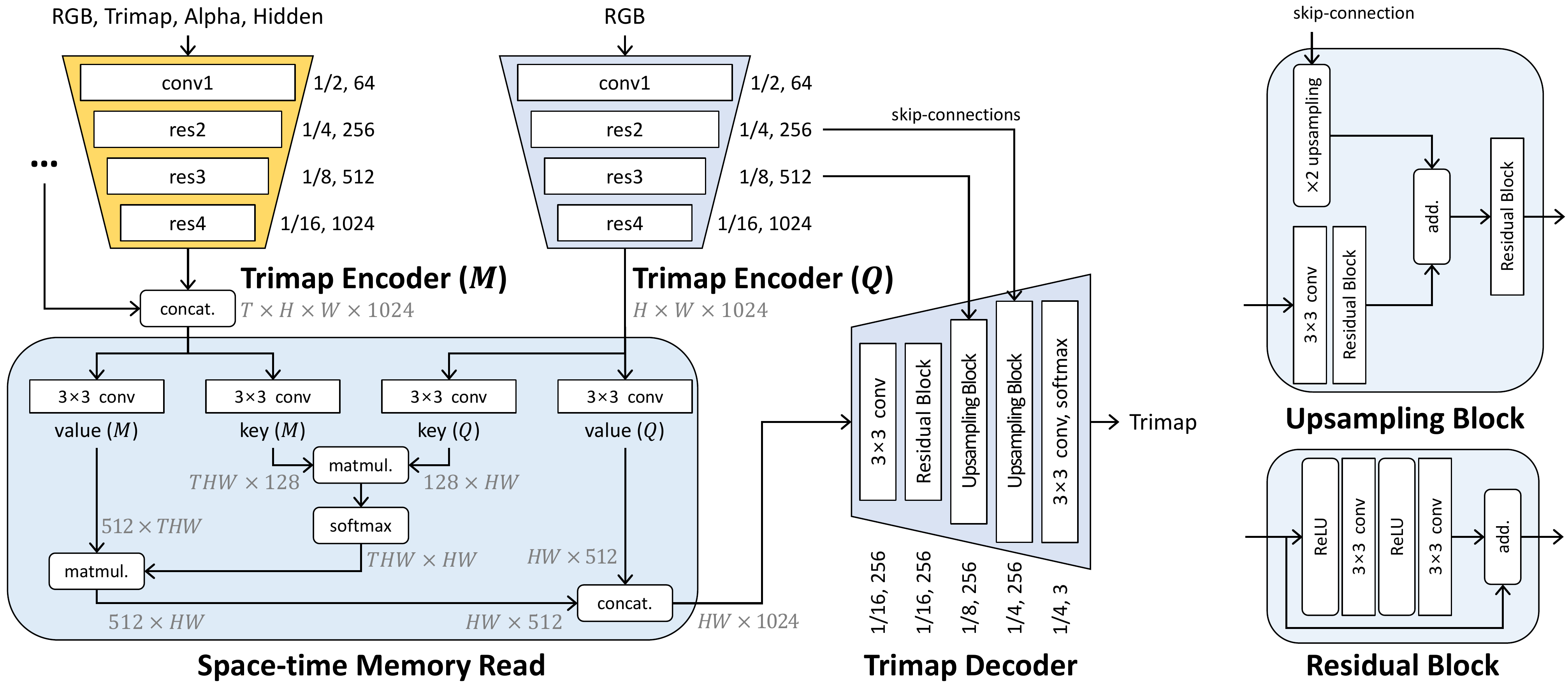}
\caption{
A detailed illustration of the trimap propagation network. Next to each block in the figure, the relative spatial scale and channel dimension of the output are notated.
\vspace{-5mm}
}
\label{fig:structure_details_trimap}
\end{figure}

\paragraph{Trimap propagation network.}~\cref{fig:structure_details_trimap} shows a detailed architecture of the trimap propagation network.
The architecture is based on STM~\cite{Oh_2019_ICCV}.
We employed two independent ResNet50~\cite{b23} encoders to embed memory and query.
Here, the last layer (\texttt{res5}) is omitted to extract fine-scale features.
The extracted memory and query features are embedded into keys and values via four independent $3 \times 3$ convolutional layers.
Using the memory key and query key, the similarity is computed via non-local matching.
Then the memory value is retrieved based on the computed similarity.
The retrieved memory value and query value are concatenated along the channel dimension, and it is fed to the trimap decoder.
In the trimap decoder, several residual blocks~\cite{he2016identity} and upsampling blocks~\cite{pinheiro2016learning,wug2018fast} are employed.
Finally, the propagated trimap is output from the trimap decoder.

\begin{figure}
\centering
\vspace{-5mm}
\includegraphics[width=1.\linewidth]{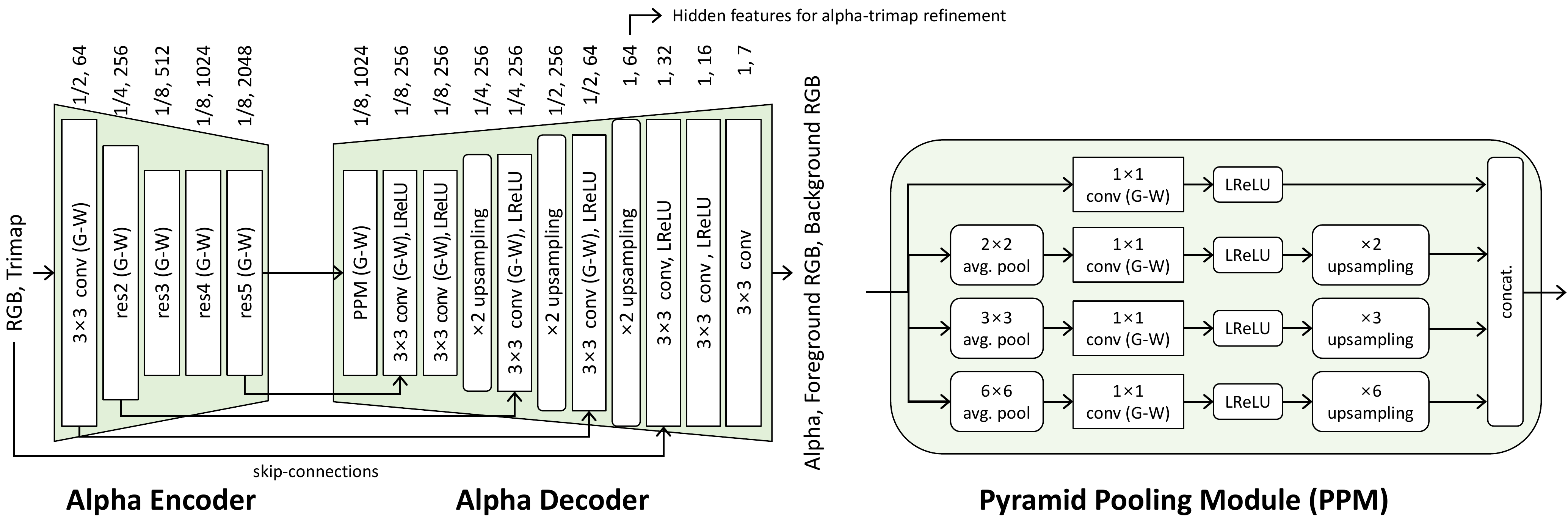}
\caption{
A detailed illustration of the alpha prediction network. In each block, (G-W) indicates Group Normalization~\cite{wu2018group} with Weight Standardization~\cite{qiao2019weight} is used. LReLU denotes Leaky ReLU~\cite{maas2013rectifier} with a negative slope of 0.01.
\vspace{-5mm}
}
\label{fig:structure_details_alpha}
\end{figure}

\paragraph{Alpha prediction network.}~A detailed implementation of the alpha prediction network is given in~\cref{fig:structure_details_alpha}.
We follow the architecture of FBA~\cite{forte2020f}.
The ResNet50~\cite{b23} with Group Normalization~\cite{wu2018group} and Weight Standardization~\cite{qiao2019weight} is used for the alpha encoder.
The alpha encoder takes an RGB frame and (either a generated or user-provided) trimap.
The three channels of the trimap are encoded into eight channels that are one channel for softmax probability of the foreground mask, one channel for softmax probability of the background mask, and six channels for three different scales of Gaussian blurs of the foreground and background masks~\cite{le2018interactive}.
In the encoder structure, the striding in the last two layers (\texttt{res4} and \texttt{res5}) is removed and the dilations of 2 and 4 are included, respectively~\cite{lutz2018alphagan}.
The alpha decoder takes the resulting pyramidal features of the alpha encoder.
In the alpha decoder, Pyramid Pooling Module (PPM)~\cite{zhao2017pyramid} is employed to increase the receptive field of the fine-scale feature.
And then, several convolutional layers, leaky ReLU~\cite{maas2013rectifier}, and bilinear upsampling are followed.
Finally, one channel of the alpha matte, three channels of the foreground RGB, three channels of the background RGB, and 64 channels of the hidden features are output from the alpha decoder.

\begin{figure}
\centering
\includegraphics[width=.8\linewidth]{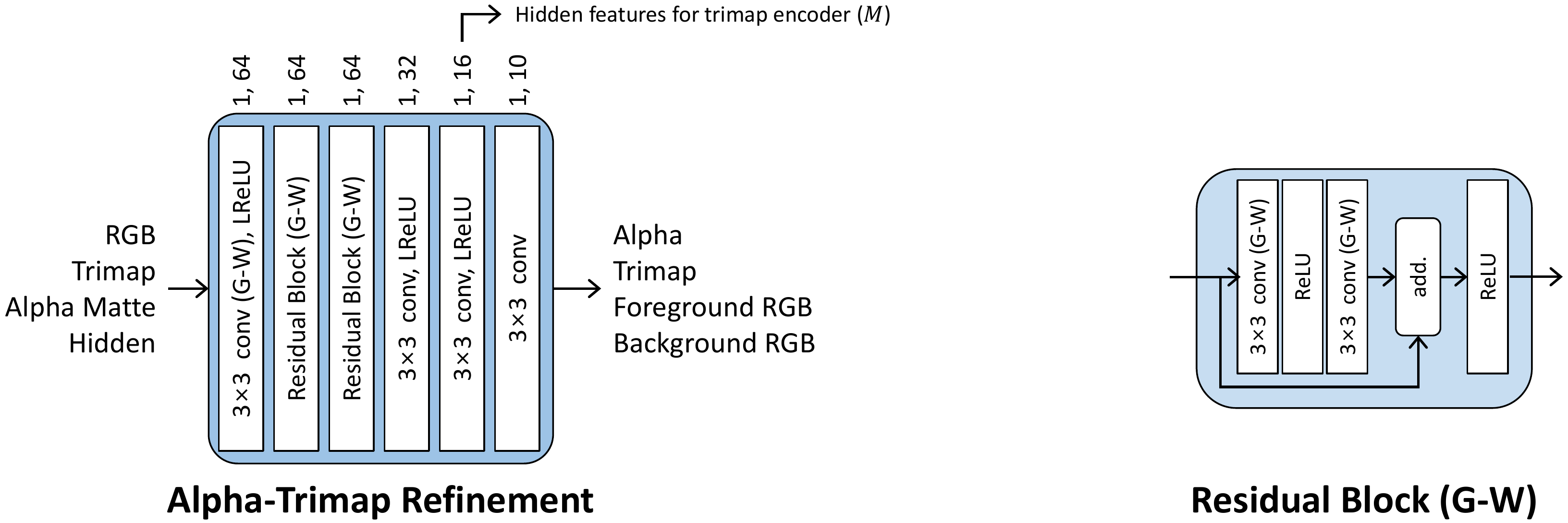}
\caption{
A detailed illustration of the alpha-trimap refinement module. 
\vspace{-5mm}
}
\label{fig:structure_details_refine}
\end{figure}

\paragraph{Alpha-trimap refinement module.}~We illustrate a detailed implementation of the alpha-trimap refinement module in~\cref{fig:structure_details_refine}.
The module takes an RGB frame, trimap, predicted alpha matte, and hidden feature which is extracted from the alpha decoder.
We employed two light-weight residual blocks with Group Normalization~\cite{wu2018group} and Weight Standardization~\cite{qiao2019weight}.
The outputs are one channel of the refined alpha matte, three channels of the trimap, three channels of the foreground RGB, three channels of the background RGB, and 16 channels of the hidden features.
All the outputs in the module will be used for the input of the trimap memory encoder.

\section{Loss Functions}
\label{sec:loss_functions}
For each predicted trimap, alpha matte, foreground RGB, and background RGB, we leverage several loss functions.
In summary, we used cross-entropy loss for the predicted trimaps, as used in STM~\cite{Oh_2019_ICCV}, and we used image matting losses used in FBA~\cite{forte2020f} and temporal coherence loss~\cite{sun2021deep} for predicted alpha mattes, foreground RGB colors, and background RGB colors.
In what follows, the specific definition of each loss function is described.

\paragraph{Trimap.}~We use the cross-entropy loss for propagated trimap ($\mathcal{L}^{tri}$) as follows:
\begin{equation}
\mathcal{L}^{tri} = { y^{tri}_{t,i} \mathrm{log} (p^{tri}_{t,i}) }
\label{eq:loss1}
\end{equation}
where $t$ and $i$ indicate time and spatial pixel index, respectively; $y^{tri}$ and $p^{tri}$ are GT trimap and propagated trimap, respectively.
The loss for refined trimap ($\widehat{\mathcal{L}}^{tri}$) is computed by simply replacing $p^{tri}$ in~\cref{eq:loss1} with refined trimap $\widehat{p}^{tri}$.
The total loss for the propagated and refined trimap is
\begin{equation}
\mathcal{L}^{tri}_{total} = \sum\limits_{t = 1}^T {\sum\limits_{i} { \mathcal{L}^{tri}}} + \sum\limits_{t = 0}^T {\sum\limits_{i} {\widehat{\mathcal{L}}^{tri} }}
\label{eq:loss2}
\end{equation}
where the reference frame (where the GT trimap is provided as the input) is given at $t=0$.

\paragraph{Alpha matte.}~With the GT alpha matte $y^{\alpha}$, the predicted alpha matte $p^{\alpha}$ extracted from the alpha decoder, input RGB frame $I$, GT foreground RGB $F$, and GT background RGB $B$, we compute the $L1$ loss ($\mathcal{L}^{\alpha}_{L1}$), compositional loss ($\mathcal{L}^{\alpha}_{comp}$)~\cite{xu2017deep}, Laplacian pyramid loss ($\mathcal{L}^{\alpha}_{lap}$)~\cite{hou2019context}, gradient loss ($\mathcal{L}^{\alpha}_{grad}$)~\cite{tang2019learning}, and temporal coherence loss ($\mathcal{L}^{\alpha}_{tc}$)~\cite{sun2021deep} as follows:
\begin{equation}
\mathcal{L}^{\alpha}_{L1} ={{{\left\| {y_{t,i}^\alpha  - p_{t,i}^\alpha } \right\|}_1}},
\label{eq:loss3}
\end{equation}
\begin{equation}
\mathcal{L}^{\alpha}_{comp} = {{{\left\| {{I_{t,i}} - p_{t,i}^\alpha {F_{t,i}} - (1 - p_{t,i}^\alpha ){B_{t,i}}} \right\|}_1}},
\label{eq:loss4}
\end{equation}
\begin{equation}
\mathcal{L}^{\alpha}_{lap} = {\sum\limits_{s = 1}^5 {{2^{s - 1}}{{\left\| {\mathcal{L}_{pyr}^s(y_{t,i}^\alpha ) - \mathcal{L}_{pyr}^s(p_{t,i}^\alpha )} \right\|}_1}} },
\label{eq:loss5}
\end{equation}
\begin{equation}
\mathcal{L}^{\alpha}_{grad} = {{{\left\| {\frac{{{\text{d}}y_{t,i}^\alpha }}{{{\text{d}}i}} - \frac{{{\text{d}}p_{t,i}^\alpha }}{{{\text{d}}i}}} \right\|}_1}},
\label{eq:loss6}
\end{equation}
\begin{equation}
\mathcal{L}^{\alpha}_{tc} = {{{\left\| {\frac{{{\text{d}}y_{t,i}^\alpha }}{{{\text{d}}t}} - \frac{{{\text{d}}p_{t,i}^\alpha }}{{{\text{d}}t}}} \right\|}_1}},
\label{eq:loss7}
\end{equation}
and the losses for the refined alpha matte ($\widehat{\mathcal{L}}^{\alpha}_{L1}, \widehat{\mathcal{L}}^{\alpha}_{comp}, \widehat{\mathcal{L}}^{\alpha}_{lap}, \widehat{\mathcal{L}}^{\alpha}_{grad}, \widehat{\mathcal{L}}^{\alpha}_{tc}$) are computed by replacing $p^{\alpha}$ in~\cref{eq:loss3,eq:loss4,eq:loss5,eq:loss6,eq:loss7} with refined alpha matte $\widehat{p}^{\alpha}$.
The total loss for the predicted and refined alpha matte is defined as follows:
\begin{equation}
\begin{split}
\mathcal{L}^{\alpha}_{total} = \sum\limits_{t} \sum\limits_{i} &  {{\mathcal{L}}^{\alpha}_{L1} + {\mathcal{L}}^{\alpha}_{comp} + {\mathcal{L}}^{\alpha}_{lap} + {\mathcal{L}}^{\alpha}_{grad} + {\mathcal{L}}^{\alpha}_{tc}} \\ & + {\widehat{\mathcal{L}}^{\alpha}_{L1} + \widehat{\mathcal{L}}^{\alpha}_{comp} + \widehat{\mathcal{L}}^{\alpha}_{lap} + \widehat{\mathcal{L}}^{\alpha}_{grad} + \widehat{\mathcal{L}}^{\alpha}_{tc} } .
\label{eq:loss8}
\end{split}
\end{equation}

\paragraph{Foreground and background colors.}~The foreground and background RGB colors are predicted from the alpha decoder ($p^{F}, p^{B}$) and the alpha-trimap refinement module ($\widehat{p}^{F}, \widehat{p}^{B}$).
Then, we compute the $L1$ losses ($\mathcal{L}^{FB}_{L1}, \widehat{\mathcal{L}}^{FB}_{L1}$), Laplacian losses ($\mathcal{L}^{FB}_{lap}, \widehat{\mathcal{L}}^{FB}_{lap}$), compositional losses ($\mathcal{L}^{FB}_{comp}, \widehat{\mathcal{L}}^{FB}_{comp}$),  gradient exclusion losses ($\mathcal{L}^{FB}_{excl}, \widehat{\mathcal{L}}^{FB}_{excl}$), and temporal coherence losses ($\mathcal{L}^{FB}_{tc}, \widehat{\mathcal{L}}^{FB}_{tc}$).
Here, the losses are computed only where the GT trimap's unknown regions ($\tilde{i} \in \text{Unknown Region}$).
Additionally, the losses for the predicted foreground color are not computed where the GT alpha value is 0 because the exact foreground color is not available in those regions.
Each loss function is defined as follows:
\begin{equation}
\mathcal{L}^{FB}_{L1} = {{{\left\| (y_{t,\tilde{i}}^\alpha > 0) ({{F_{t,\tilde{i}}} - p_{t,\tilde{i}}^F}) \right\|}_1} + {{\left\| {{B_{t,\tilde{i}}} - p_{t,\tilde{i}}^B} \right\|}_1}} ,
\label{eq:loss9}
\end{equation}
\begin{equation}
\mathcal{L}^{FB}_{comp} = {{{\left\| {{I_{t,\tilde{i}}} - y_{t,\tilde{i}}^\alpha p_{t,\tilde{i}}^F - (1 - y_{t,\tilde{i}}^\alpha )p_{t,\tilde{i}}^B} \right\|}_1}} ,
\label{eq:loss10}
\end{equation}
\begin{equation}
\begin{split}
\mathcal{L}^{FB}_{lap} = \sum\limits_{s = 1}^5 {2^{s - 1}} & \Big( {{\left\| {(y_{t,\tilde i}^\alpha  > 0)(\mathcal{L}_{pyr}^s({F_{t,\tilde i}}) - \mathcal{L}_{pyr}^s(p_{t,\tilde i}^F))} \right\|}_1} \\ & + {{\left\| {\mathcal{L}_{pyr}^s({B_{t,\tilde i}}) - \mathcal{L}_{pyr}^s(p_{t,\tilde i}^B)} \right\|}_1} \Big)  ,
\label{eq:loss11}
\end{split}
\end{equation}
\begin{equation}
\mathcal{L}^{FB}_{excl} = {\left\| {(y_{t,\tilde i}^\alpha  > 0)\frac{{{\text{d}}p_{t,\tilde i}^F}}{{{\text{d}}\tilde i}}} \right\|_1}{\left\| {\frac{{{\text{d}}p_{t,\tilde i}^B}}{{{\text{d}}\tilde i}}} \right\|_1} ,
\label{eq:loss12}
\end{equation}
\begin{equation}
\mathcal{L}^{FB}_{tc} = {\left\| {(y_{t,\tilde i}^\alpha  > 0)(\frac{{{\text{d}}{F_{t,\tilde i}}}}{{{\text{d}}t}} - \frac{{{\text{d}}p_{t,\tilde i}^F}}{{{\text{d}}t}})} \right\|_1} + {\left\| {\frac{{{\text{d}}{B_{t,\tilde i}}}}{{{\text{d}}t}} - \frac{{{\text{d}}p_{t,\tilde i}^B}}{{{\text{d}}t}}} \right\|_1} ,
\label{eq:loss13}
\end{equation}
and the losses for the predicted colors extracted from the alpha-trimap refinement module ($\widehat{\mathcal{L}}^{FB}_{L1}, \widehat{\mathcal{L}}^{FB}_{lap}, \widehat{\mathcal{L}}^{FB}_{comp}, \widehat{\mathcal{L}}^{FB}_{excl}, \widehat{\mathcal{L}}^{FB}_{tc}$) are computed by replacing ${p}^{F}, {p}^{B}$ in~\cref{eq:loss9,eq:loss10,eq:loss11,eq:loss12,eq:loss13} with $\widehat{p}^{F}, \widehat{p}^{B}$, respectively.
The total loss for the foreground and background RGB colors is defined by
\begin{equation}
\begin{split}
\mathcal{L}^{FB}_{total} = \sum\limits_{t} \sum\limits_{\tilde{i}} & {\mathcal{L}}^{FB}_{L1} + {\mathcal{L}}^{FB}_{comp} + {\mathcal{L}}^{FB}_{lap} + {\mathcal{L}}^{FB}_{excl} + {\mathcal{L}}^{FB}_{tc} \\ & + \widehat{\mathcal{L}}^{FB}_{L1} + \widehat{\mathcal{L}}^{FB}_{comp} + \widehat{\mathcal{L}}^{FB}_{lap} + \widehat{\mathcal{L}}^{FB}_{excl} + \widehat{\mathcal{L}}^{FB}_{tc}  .
\label{eq:loss14}
\end{split}
\end{equation}

Finally, all loss functions are summarized by
\begin{equation}
\mathcal{L}_{total} = \mathcal{L}^{tri}_{total} + \mathcal{L}^{\alpha}_{total} + 0.25 \mathcal{L}^{FB}_{total} .
\label{eq:loss15}
\end{equation}

\section{Trimap Input for the Trimap Encoder}
Ideally, the hidden feature can subsume the trimap information. We study the effect of the trimap input for the trimap encoder, and the results are given in \cref{tab_ablation_trimap}. We empirically found that explicitly providing the trimap input is helpful. We conjecture that trimap input facilitates the training of the trimap propagation module under the insufficient video training dataset.

\begin{table}
\vspace{-.5cm}
\caption{Ablation on trimap input for trimap encoder.}
\label{tab_ablation_trimap}
\centering
\mytabular{
\setlength{\tabcolsep}{3pt}
\begin{tabular}{l|ccccc|ccccc}
\toprule
\multicolumn{1}{c|}{Trimap Encoder Inputs} & SSDA-V         & MSE-V         & MAD-V          & dtSSD-V        & MESSDdt-V     & SSDA           & MSE           & MAD            & dtSSD          & MESSDdt       \\
\midrule
trimap+alpha+hidden   & \textbf{54.67} & \textbf{2.61} & \textbf{13.02} & \textbf{29.87} & \textbf{1.78} & \textbf{50.51} & \textbf{8.58} & \textbf{37.16} & \textbf{28.28} & \textbf{1.63} \\
alpha+hidden          & 75.08          & 9.42          & 22.03          & 31.95          & 2.98          & 67.50          & 18.88         & 50.21          & 29.37          & 2.67          \\
hidden                & 130.54         & 31.09         & 43.84          & 32.88          & 3.84          & 77.30          & 37.70         & 69.36          & 28.92          & 2.71         \\
\bottomrule
\end{tabular}
}
\end{table}
\vspace{-1cm}

\section{Visual Analysis of the Hidden Feature}
\begin{figure}
\vspace{-.5cm}
\centering
\includegraphics[width=1.\linewidth]{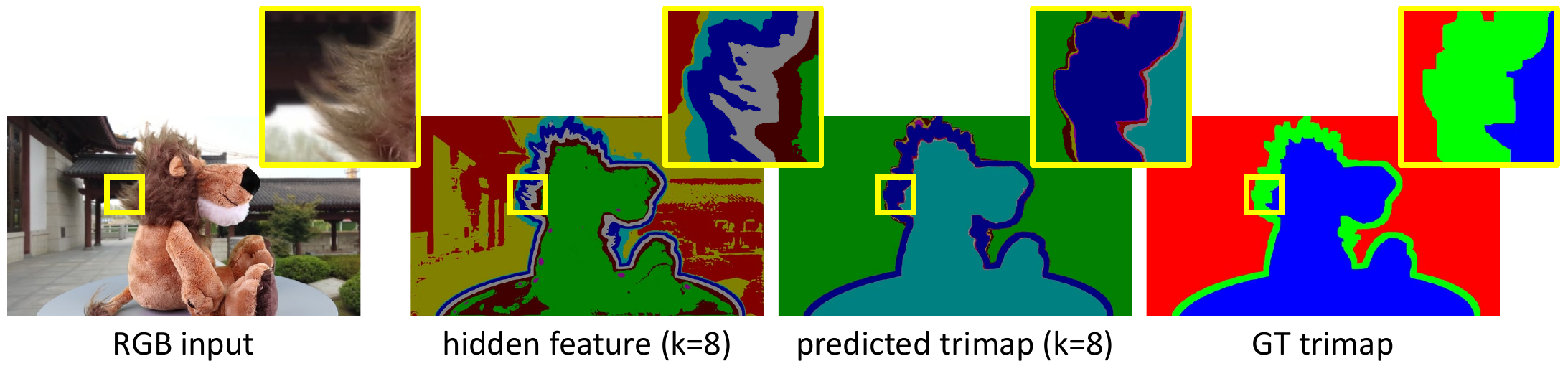}
\vspace{-0.6cm}
\caption{
Visualization of the hidden feature and trimap. For a fair comparison, we apply k-mean clustering to both hidden feature and predicted trimap.}
\label{fig:supp_visualization_of_hidden_feature}
\vspace{-.5cm}
\end{figure}

To analyze what information is contained in the hidden features, we visualize the learned hidden feature through k-means clustering (k=8) in \cref{fig:supp_visualization_of_hidden_feature}. For a fair comparison, we also apply k-means clustering to the predicted trimap. As shown in the figure, the hidden feature embeds more information than trimap: (1) it subdivides the unknown regions into several levels; (2) it includes semantic information in the background regions that would be helpful to estimate accurate background regions of the next frame.

\section{Effect of the Refinement Module on Trimap Estimation}
\begin{figure}
\centering
\includegraphics[width=1.\linewidth]{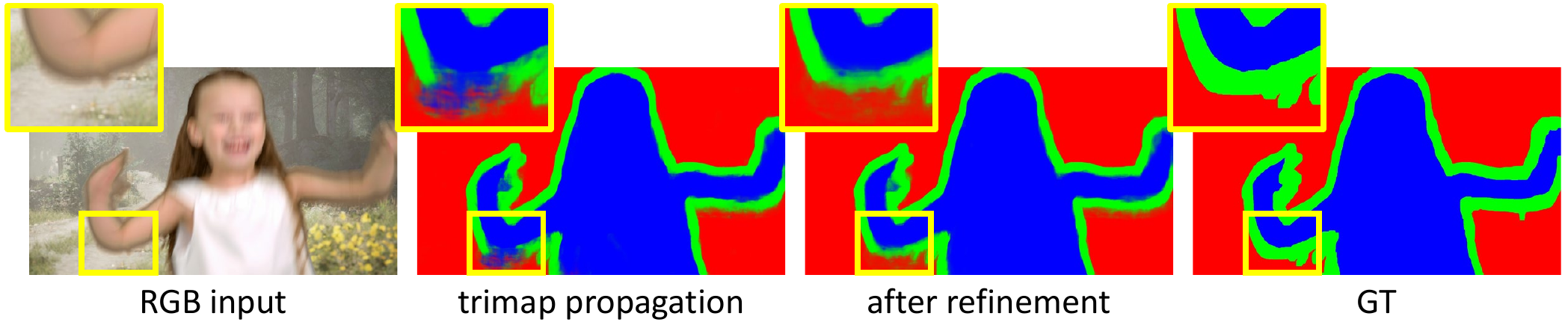}
\vspace{-0.6cm}
\caption{
Effect of the trimap refinement.}
\label{fig:supp_effect_of_refinement_module}
\end{figure}

To clearly show the effects of the refinement module, we measure a trimap performance of the output from the trimap propagation module in OTVM. The result is given in \cref{tab_trimap_performance}. We further show the effect of the refinement module qualitatively in \cref{fig:supp_effect_of_refinement_module}. In the figure, the trimap propagation fails in the zoomed region due to motion blur, while the refinement module corrects it.

\begin{table}
\caption{Trimap performance. ``-T'' is presented to estimate trimap quality and denotes that the unknown region in GT trimap has been modified (see Sec.~4.2 in the main paper).}
\label{tab_trimap_performance}
\centering
\mytabular{
\setlength{\tabcolsep}{10pt}
\begin{tabular}{l|ccc}
\toprule
\multicolumn{1}{c|}{Method}                         & Precision-T    & Recall-T       & Average        \\
\midrule
Decoupled STM~\cite{sun2021deep,zhang2021attention} & 96.98          & 93.58          & 95.28          \\
OTVM (from trimap propagation)                      & 98.00          & 95.29          & 96.65          \\
OTVM (from refinement module)                       & \textbf{98.17} & \textbf{95.92} & \textbf{97.05}\\
\bottomrule
\end{tabular}
}
\end{table}

\vspace{-.5cm}
\section{Runtime and GPU Memory Consumption}
\begin{figure}
\centering
\vspace{-.5cm}
\includegraphics[width=.5\linewidth]{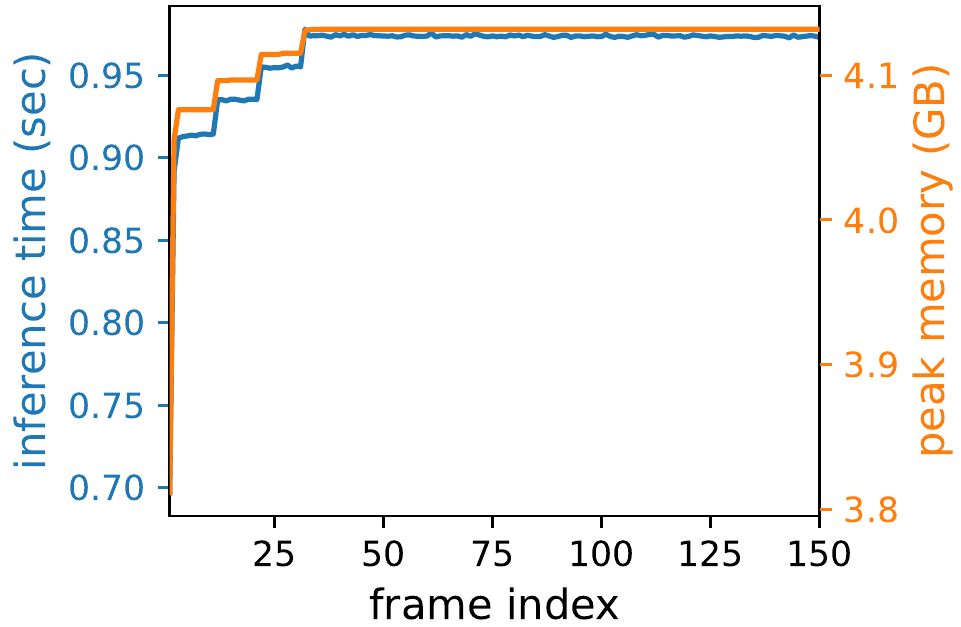}
\vspace{-0.3cm}
\caption{
Inference time and peak GPU memory.}
\label{fig:supp_time_memory_graph}
\vspace{-1cm}
\end{figure}

In \cref{fig:supp_time_memory_graph}, we show the inference time and memory consumption at each frame. We used high-resolution (1920$\times$1080) video and tested with one NVIDIA GeForce 1080 Ti GPU. As shown in the figure, OTVM slows down and consumes more memory for every 10 frames because we store the intermediate frames for trimap propagation. Instead of keeping all intermediate memory frames, we avoid this from the 30th frame by limiting our maximum memory size to 5 frames and storing only the first frame, the last frame, and up to three latest intermediate frames to the memory.

\section{Temporal Stability}
\begin{figure}
\centering
\includegraphics[width=1\linewidth]{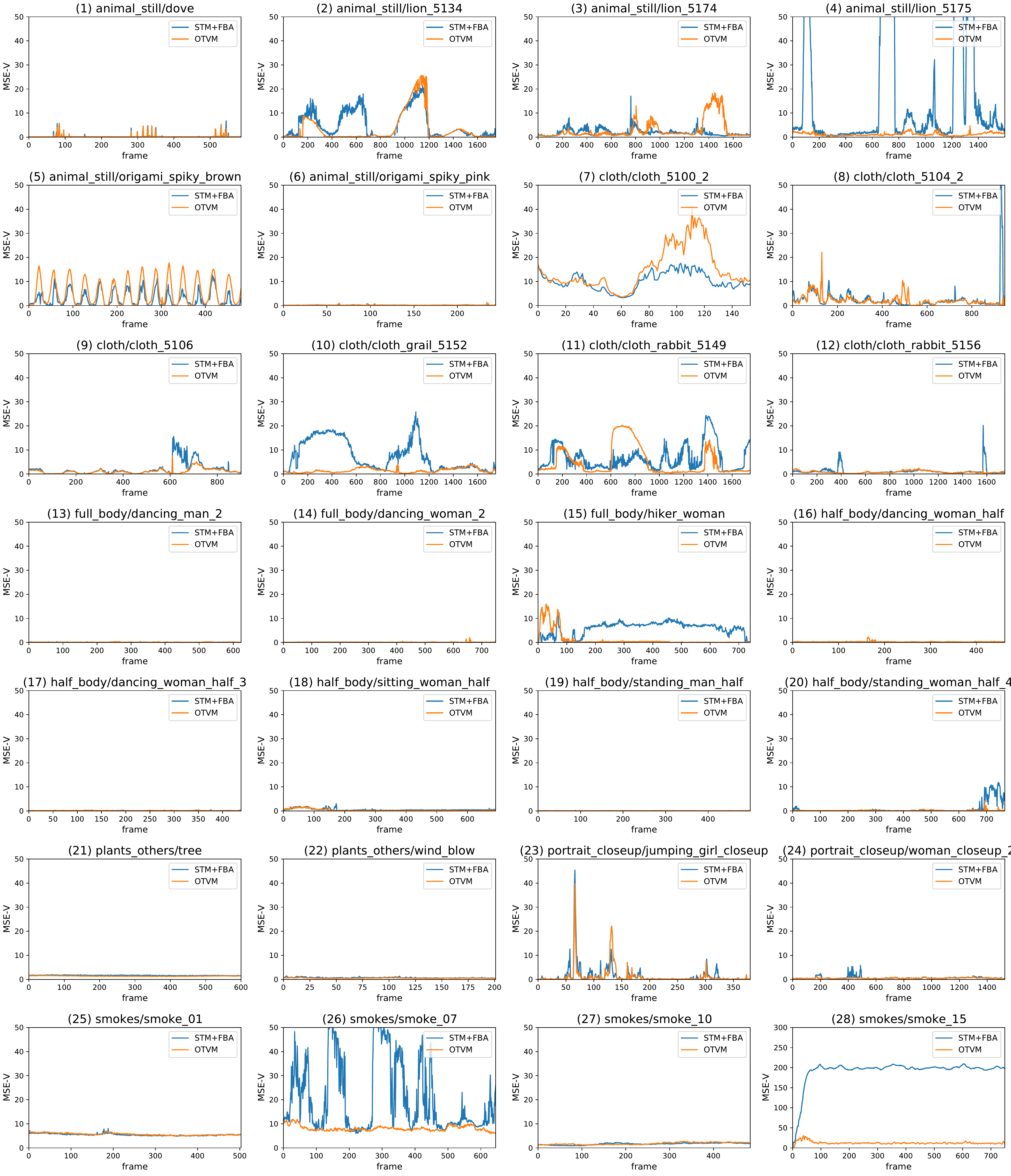}
\vspace{-.7cm}
\caption{
Per frame comparison. The results are obtained on VideoMatting108 validation set with medium trimap setting. We plot all sequences in the validation set. Note that the lower MSE-V is better.}
\vspace{-.8cm}
\label{fig:results_over_time}
\end{figure}

To demonstrate the superiority of OTVM in terms of temporal stability, we show per frame comparison in \cref{fig:results_over_time}. In the figure, we did not cherry-pick the results and show results in all sequences of VideoMatting108 validation set. As shown in \cref{fig:results_over_time}, decoupled approach, \ie, STM+FBA, is extremely unstable in sequences (4), (10), (15), (26), and (28). In contrast, OTVM achieves temporally stable results in most sequences.

\clearpage

\section{More Quantitative Results}
To encourage comparison for future works, we present additional results by measuring errors in a different way from the tables in the main paper.
Specifically, we computed errors on the full-frame to capture the errors that occurred by inaccurate trimap propagation, and we denoted it with ``-V'' in Table~1 of the main paper.
We re-measure by computing errors only on the unknown regions according to the official metric and report in \cref{tab1_supp}.
In contrast to those, Table~2 in the main paper is computed errors only on the unknown regions for fair comparisons with previous works.
We re-measure of our implemented methods by computing errors on the full-frame and report in \cref{table:comparison_supp}.

Furthermore, following~\cite{zhang2021attention}, we show results with narrow and wide trimap settings in \cref{table:comparison_supp_V108,table:comparison_supp_V1082}. As shown in the tables, OTVM always outperforms the state-of-the-art methods in any trimap settings.

\begin{table}
\caption{Analysis experiments on VideoMatting108 validation set. For all experiments, we use 1-trimap setting where GT trimap is given only at the first frame.}
\label{tab1_supp}
\centering
\setlength{\tabcolsep}{6pt}
\vspace{-3mm}\mytabular{(a) \textbf{Joint modeling}.}\\
\mytabular{
\setlength{\tabcolsep}{7.1pt}
\begin{tabular}{cc|ccccc}
\toprule
Model                    & Training method & SSDA          & MSE          & MAD           & dtSSD         & MESSDdt        \\
\midrule
\multirow{2}{*}{STM+FBA} & decoupled       & 70.63 & \textbf{20.18} & \textbf{51.20} & 31.00 & 2.86          \\
                         & joint      & \textbf{68.93} & 21.10 & 51.57 & \textbf{28.18} & \textbf{2.56}          \\
                         \bottomrule
\end{tabular}
}
\\ \vspace{2mm}\mytabular{(b) \textbf{Stage-wise training}.}\\
\mytabular{
\setlength{\tabcolsep}{8.1pt}
\begin{tabular}{cc|ccccc}
\toprule
Model                    & Training method & SSDA          & MSE          & MAD           & dtSSD         & MESSDdt        \\
\midrule
\multirow{2}{*}{OTVM}    & joint      & 66.08 & 17.71 & 47.84 & \textbf{28.28} & 2.53          \\
                         & joint + stage-wise      & \textbf{50.51} & \textbf{8.58} & \textbf{37.16} & \textbf{28.28} & \textbf{1.63}\\
                         \bottomrule
\end{tabular}
}
\\ \vspace{2mm}\mytabular{(c) \textbf{Ablation on training stages}.
}\\
\mytabular{
\setlength{\tabcolsep}{5.3pt}
\begin{tabular}{cccc|ccccc}
\toprule
\multicolumn{4}{c|}{Train stages}            & \multirow{2}{*}{SSDA} & \multirow{2}{*}{MSE} & \multirow{2}{*}{MAD} & \multirow{2}{*}{dtSSD} & \multirow{2}{*}{MESSDdt} \\
Stage 1    & Stage 2    & Stage 3    & Stage 4    &                       &                      &                      &                        &                        \\
\midrule
           &            &            & \checkmark & 73.66                 & 24.30                & 55.81                & 30.39                  & 2.72                   \\
\checkmark &            &            & \checkmark & 66.81                 & 18.24                & 48.91                & 28.20                  & 2.58                   \\
\checkmark & \checkmark &            & \checkmark & 67.96                 & 18.63                & 50.33                & 28.88                  & 2.66                   \\
\checkmark & \checkmark & \checkmark & \checkmark & \textbf{50.51}        & \textbf{8.58}        & \textbf{37.16}       & \textbf{28.28}         & \textbf{1.63}         \\
\bottomrule
\end{tabular}
}
\\ \vspace{2mm}\mytabular{(d) \textbf{Ablation on modules}.
} \\
\mytabular{
\setlength{\tabcolsep}{2.5pt}
\begin{tabular}{cccc|cccccc}
\toprule
\multicolumn{2}{c}{\begin{tabular}[c]{@{}c@{}}Refinement module\\ (output)\end{tabular}} & \multicolumn{2}{c|}{\begin{tabular}[c]{@{}c@{}}Trimap module\\ (input)\end{tabular}} & \multirow{2}{*}{SSDA} & \multirow{2}{*}{MSE} & \multirow{2}{*}{MAD} & \multirow{2}{*}{dtSSD} & \multirow{2}{*}{MESSDdt} & \multirow{2}{*}{{\begin{tabular}[c]{@{}c@{}}Time\\(sec/frame)\end{tabular}}} \\
Alpha                                  & Trimap                                & Alpha                                & Hidden                               &                       &                      &                      &                        & &                       \\
\midrule
                                       &                                       &                                      &                                      & 69.64                 & 29.38                & 59.46                & 27.90                  & 2.53  & 0.799                 \\
\checkmark                             &                                       &                                      &                                      & 70.95                 & 27.68                & 57.59                & 28.52                  & 2.57 & 0.951                  \\
                                       & \checkmark                            &                                      &                                      & 69.73                 & 22.41                & 53.05                & 28.22                  & 2.61 & 0.946                  \\
\checkmark                             & \checkmark                            &                                      &                                      & 66.96                 & 17.96                & 48.24                & 28.62                  & 2.57 & 0.952                   \\
\checkmark                             & \checkmark                            & \checkmark                           &                                      & 52.64                 & 14.16                & 42.89                & 27.78                  & \textbf{1.59} & 0.955          \\
\checkmark                             & \checkmark                            & \checkmark                           & \checkmark                           & \textbf{50.51}        & \textbf{8.58}        & \textbf{37.16}       & \textbf{28.28}         & 1.63 & 0.964                 \\
\bottomrule
\end{tabular}
}
\\\vspace{2mm}\mytabular{(e) \textbf{Different image matting backbones}.} \\
\mytabular{
\setlength{\tabcolsep}{9.1pt}
\begin{tabular}{cl|ccccc}
\toprule
Backbone                                  & Model               & SSDA          & MSE           & MAD           & dtSSD         & MESSDdt        \\
\midrule
\multirow{2}{*}{DIM~\cite{xu2017deep}}    & STM+DIM  & 88.49 & \textbf{25.36} & 66.98 & 41.64 & 4.40          \\
                                          & OTVM                 & \textbf{86.83} & 25.66 & \textbf{64.92} & \textbf{37.47} & \textbf{3.88} \\
                                          \midrule
\multirow{2}{*}{GCA~\cite{li2020natural}} & STM+GCA  & 84.82 & 24.38 & 66.01 & 36.14 & 3.60          \\
                                          & OTVM                 & \textbf{77.75} & \textbf{23.24} & \textbf{60.13} & \textbf{33.47} & \textbf{3.12} \\
                                          \bottomrule
\end{tabular}
}
\end{table}

\begin{table}
\setlength{\tabcolsep}{7pt}
\caption{Comparison with state-of-the-art methods on public benchmarks. The trimap setting indicates how many GT trimaps are given as input, \ie, ``full-trimap" for all frames, ``20/40-frame" for every 20/40th frames, ``1-trimap" for only at the first frame.}
\label{table:comparison_supp}
\centering
\vspace{-0.2cm} \mytabular{
\begin{tabular}{c}
(a) \textbf{Comparison on VideoMatting108 validation set}.
In this experiment, we use the medium trimap setting.\\
$^\dagger$ denotes our reproduced results using our training setup.
\end{tabular}
} \\
\mytabular{
\begin{tabular}{cl|ccccc}
\toprule
Trimap   Setting             & Methods                                          & SSDA-V           & MSE-V           & MAD-V            & dtSSD-V          & MESSDdt-V         \\
\midrule
\multirow{2}{*}{full-trimap} & TCVOM (GCA)~\cite{zhang2021attention}                   & 50.41 & 2.14 & 12.80 &  27.28 & 1.48          \\
                             & TCVOM   (FBA)$^\dagger$~\cite{zhang2021attention}       & \textbf{39.76} & \textbf{1.41} & \textbf{10.56} & \textbf{22.93} & \textbf{1.06} \\
                             \midrule
\multirow{4}{*}{1-trimap}    & STM + TCVOM (GCA)~\cite{zhang2021attention}             & 89.93 & 11.04 & 24.09 & 37.51 & 3.46          \\
                             & STM + TCVOM   (FBA)$^\dagger$~\cite{zhang2021attention} & 81.97 & 10.24 & 22.22 & 34.68 & 3.17          \\
                             & STM +   FBA$^\dagger$~\cite{forte2020f}          & 83.61 & 10.62 & 22.12 & 36.31 & 3.45          \\
                             & OTVM                                             & \textbf{54.67} & \textbf{2.61} & \textbf{13.02} & \textbf{29.87} & \textbf{1.78} \\
                             \bottomrule
\end{tabular}
}
\\ \vspace{2mm} \mytabular{(b) \textbf{Comparison on DVM validation set}.
}\\
\mytabular{
\begin{tabular}{cl|cccccc}
\toprule
Trimap   Setting & Methods & SAD-V & MSE-V & Grad-V & Conn-V & dtSSD-V & MESSDdt-V \\
\midrule
20-frame         & OTVM    & 40.61 & 0.005 & 19.55  & 38.81  & 27.53   & 0.08    \\
\midrule
40-frame         & OTVM    & 41.28 & 0.005 & 19.74  & 39.44  & 27.69   & 0.08    \\
\midrule
1-trimap         & OTVM    & 50.64 & 0.007 & 20.88  & 48.45  & 27.75   & 0.10   \\
\bottomrule
\end{tabular}
}
\end{table}

\begin{table}
\setlength{\tabcolsep}{7pt}
\caption{Comparison on VideoMatting108 validation set. We compute the error in all regions of the trimap}
\label{table:comparison_supp_V108}
\centering
\vspace{-0.2cm} \mytabular{
\begin{tabular}{c}
(a) \textbf{Narrow trimap setting}.
\end{tabular}
} \\
\mytabular{
\begin{tabular}{cl|ccccc}
\toprule
Trimap   Setting             & Methods                                                 & SSDA-V         & MSE-V         & MAD-V          & dtSSD-V        & MESSDdt-V       \\
\midrule
\multirow{2}{*}{full-trimap} & TCVOM (GCA)~\cite{zhang2021attention}                   & 45.39          & 1.88          & 12.02          & 24.37          & 1.28          \\
                             & TCVOM   (FBA)$^\dagger$~\cite{zhang2021attention}       & \textbf{37.03} & \textbf{1.24} & \textbf{9.87}  & \textbf{21.09} & \textbf{0.93} \\
                             \midrule
\multirow{4}{*}{1-trimap}    & STM + TCVOM (GCA)~\cite{zhang2021attention}             & 86.73          & 10.87         & 23.40          & 35.60          & 3.29          \\
                             & STM + TCVOM   (FBA)$^\dagger$~\cite{zhang2021attention} & 81.02          & 10.16         & 21.72          & 33.48          & 3.07          \\
                             & STM +   FBA$^\dagger$~\cite{forte2020f}                 & 80.91          & 10.21         & 21.28          & 34.45          & 3.15          \\
                             & OTVM                                                    & \textbf{57.58} & \textbf{2.91} & \textbf{13.13} & \textbf{29.24} & \textbf{1.72}\\
                             \bottomrule
\end{tabular}
}
\\ \vspace{2mm} \mytabular{(b) \textbf{Wide trimap setting}.
}\\
\mytabular{
\begin{tabular}{cl|ccccc}
\toprule
Trimap   Setting             & Methods                                                 & SSDA-V         & MSE-V         & MAD-V          & dtSSD-V        & MESSDdt-V     \\
\midrule
\multirow{2}{*}{full-trimap} & TCVOM (GCA)~\cite{zhang2021attention}                   & 54.35          & 2.38          & 13.56          & 29.60          & 1.69          \\
                             & TCVOM   (FBA)$^\dagger$~\cite{zhang2021attention}       & \textbf{46.52} & \textbf{1.79} & \textbf{12.15} & \textbf{26.60} & \textbf{1.35} \\
                             \midrule
\multirow{4}{*}{1-trimap}    & STM + TCVOM (GCA)~\cite{zhang2021attention}             & 97.94          & 12.08         & 26.21          & 39.21          & 3.78          \\
                             & STM + TCVOM   (FBA)$^\dagger$~\cite{zhang2021attention} & 88.97          & 11.12         & 24.54          & 36.67          & 3.51          \\
                             & STM +   FBA$^\dagger$~\cite{forte2020f}                 & 90.78          & 11.48         & 24.21          & 38.62          & 3.89          \\
                             & OTVM                                                    & \textbf{74.92} & \textbf{7.25} & \textbf{18.25} & \textbf{31.44} & \textbf{2.00}\\
                             \bottomrule
\end{tabular}
}
\end{table}

\begin{table}
\setlength{\tabcolsep}{7pt}
\caption{Comparison on VideoMatting108 validation set. We compute the error in unknown regions of the trimap}
\label{table:comparison_supp_V1082}
\centering
\vspace{-0.2cm} \mytabular{
\begin{tabular}{c}
(a) \textbf{Narrow trimap setting}.
\end{tabular}
} \\
\mytabular{
\begin{tabular}{cl|ccccc}
\toprule
Trimap   Setting             & Methods                                                 & SSDA           & MSE            & MAD            & dtSSD          & MSDdt         \\
\midrule
\multirow{5}{*}{full-trimap} & DIM~\cite{xu2017deep}                                   & 56.40          & 10.46          & 51.76          & 31.77          & 2.56          \\
                             & IndexNet~\cite{lu2019indices}                           & 52.75          & 9.78           & 50.90          & 29.49          & 1.97          \\
                             & GCA~\cite{li2020natural}                                & 49.99          & 8.32           & 46.86          & 27.91          & 1.80          \\
                             & TCVOM   (GCA)~\cite{zhang2021attention}                 & 45.39          & 7.30           & 44.01          & 24.37          & 1.28          \\
                             & TCVOM   (FBA)$^\dagger$~\cite{zhang2021attention}       & \textbf{37.03} & \textbf{4.53}  & \textbf{34.57} & \textbf{21.09} & \textbf{0.93} \\
                             \midrule
\multirow{4}{*}{1-trimap}    & STM + TCVOM (GCA)~\cite{zhang2021attention}             & 72.29          & 23.41          & 66.22          & 29.87          & 2.78          \\
                             & STM + TCVOM   (FBA)$^\dagger$~\cite{zhang2021attention} & 66.94          & 21.52          & 60.52          & 28.05          & 2.59          \\
                             & STM +   FBA$^\dagger$~\cite{forte2020f}                 & 66.79          & 21.76          & 60.37          & 29.01          & 2.65          \\
                             & OTVM                                                    & \textbf{48.83} & \textbf{12.34} & \textbf{48.50} & \textbf{26.88} & \textbf{1.52}\\
                             \bottomrule
\end{tabular}
}
\\ \vspace{2mm} \mytabular{(b) \textbf{Wide trimap setting}.
}\\
\mytabular{
\begin{tabular}{cl|ccccc}
\toprule
Trimap   Setting             & Methods                                                 & SSDA           & MSE            & MAD            & dtSSD          & MSDdt         \\
\midrule
\multirow{5}{*}{full-trimap} & DIM~\cite{xu2017deep}                                   & 67.15          & 10.25          & 41.88          & 37.64          & 3.21          \\
                             & IndexNet~\cite{lu2019indices}                           & 64.49          & 9.73           & 41.22          & 36.39          & 2.73          \\
                             & GCA~\cite{li2020natural}                                & 60.69          & 8.41           & 38.59          & 34.83          & 2.50          \\
                             & TCVOM   (GCA)~\cite{zhang2021attention}                 & 54.35          & 6.98           & 34.81          & 29.60          & 1.69          \\
                             & TCVOM   (FBA)$^\dagger$~\cite{zhang2021attention}       & \textbf{46.52} & \textbf{4.84}  & \textbf{30.45} & \textbf{26.60} & \textbf{1.35} \\
                             \midrule
\multirow{4}{*}{1-trimap}    & STM + TCVOM (GCA)~\cite{zhang2021attention}             & 85.05          & 24.31          & 56.65          & 34.61          & 3.22          \\
                             & STM + TCVOM   (FBA)$^\dagger$~\cite{zhang2021attention} & 77.59          & \textbf{22.17} & 52.62          & 32.51          & 3.01          \\
                             & STM +   FBA$^\dagger$~\cite{forte2020f}                 & 78.57          & 22.71          & 52.44          & 33.98          & 3.18          \\
                             & OTVM                                                    & \textbf{61.69} & 24.84          & \textbf{50.85} & \textbf{29.93} & \textbf{1.83}\\
                             \bottomrule
\end{tabular}
}
\end{table}

\section{More Qualitative Results}
\begin{figure}
\centering
\includegraphics[width=1.\linewidth]{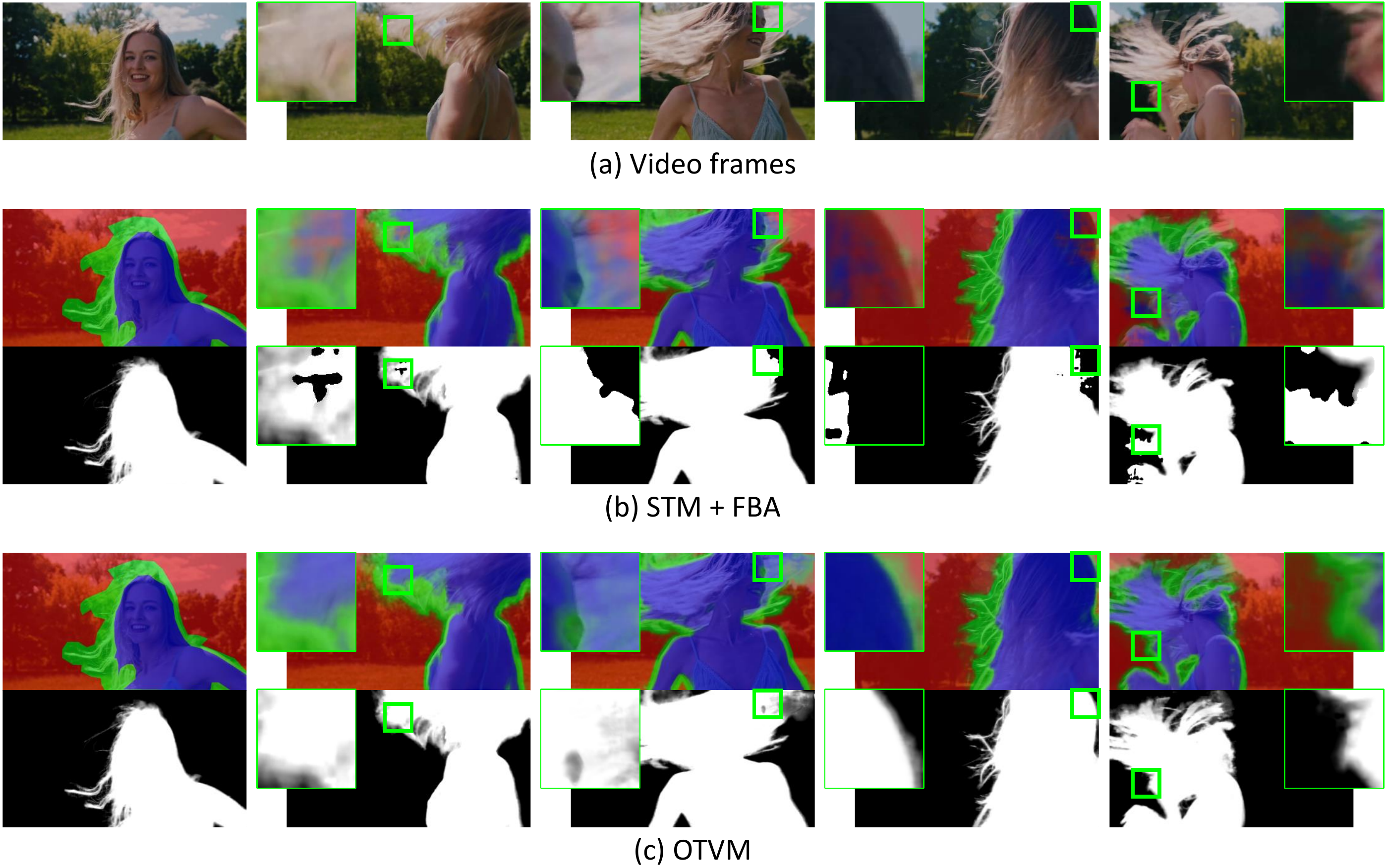}
\caption{
Qualitative results on a real-world video.}
\label{fig:supp_qualitative_results_real_video_1}
\end{figure}

\begin{figure}
\centering
\includegraphics[width=1.\linewidth]{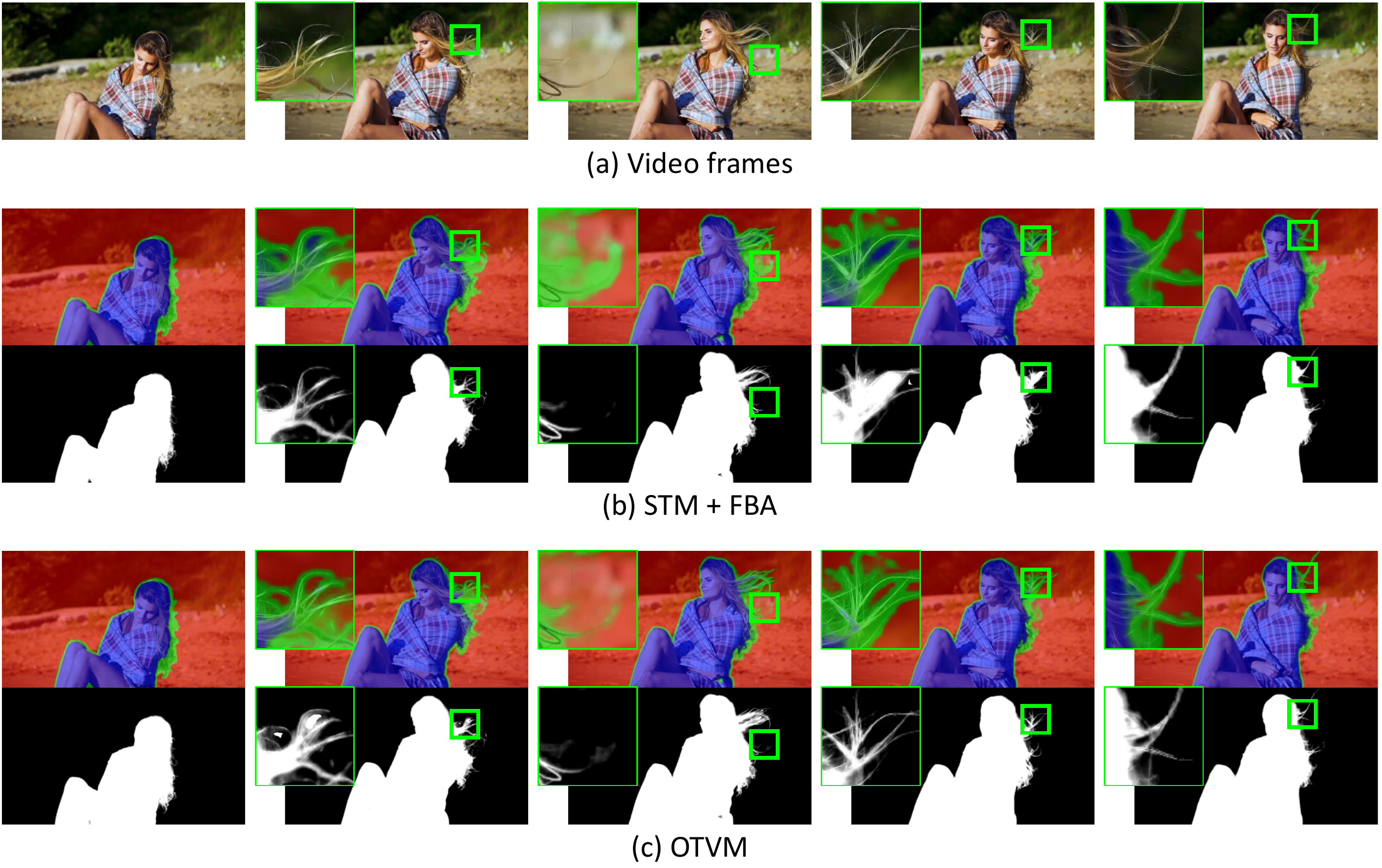}
\caption{
Qualitative results on a real-world video.}
\label{fig:supp_qualitative_results_real_video_2}
\end{figure}

\begin{figure}
\centering
\includegraphics[width=1.\linewidth]{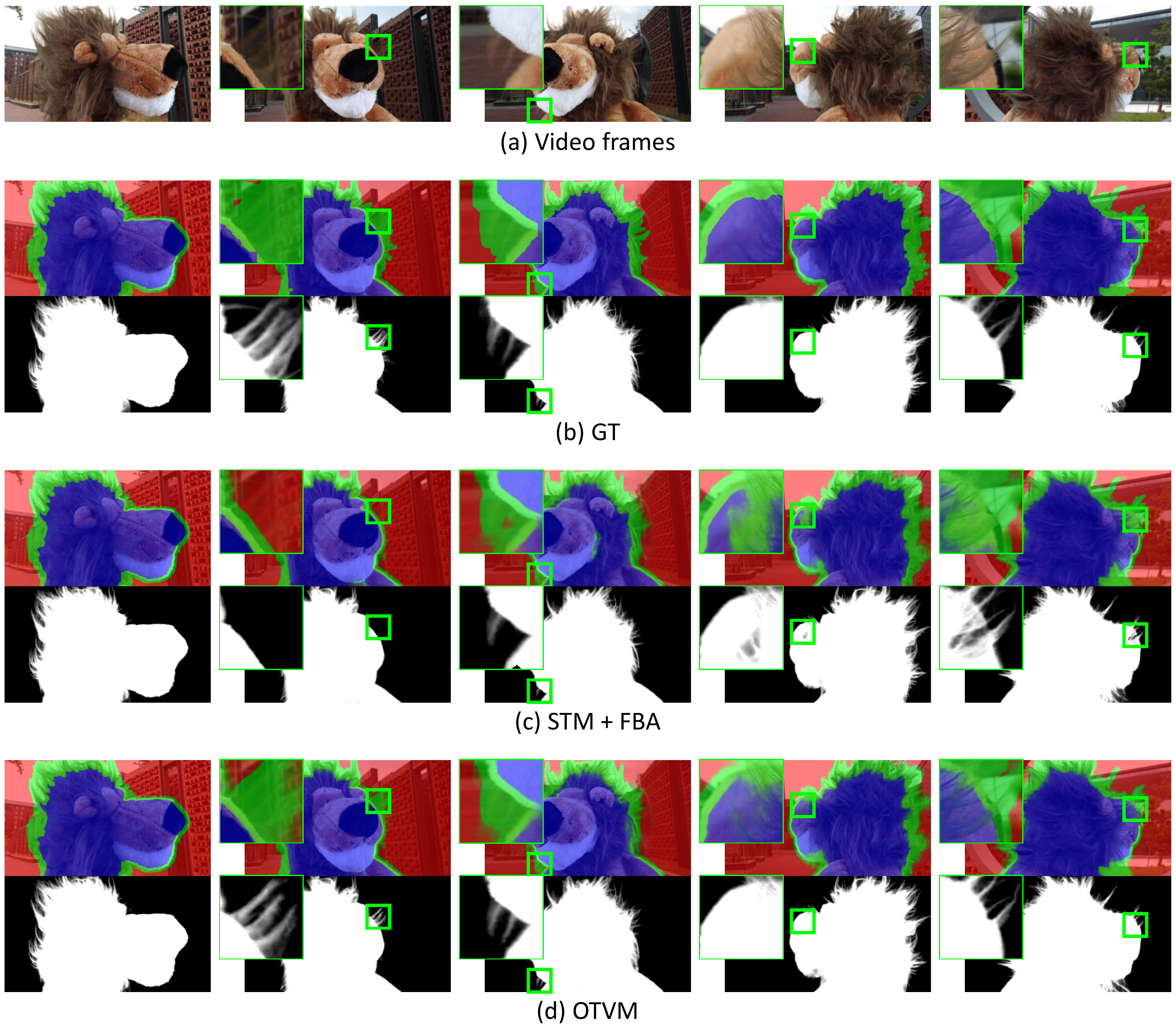}
\caption{
Qualitative results on VideoMatting108 validation set.}
\label{fig:supp_qualitative_results_V108_1}
\end{figure}

\begin{figure}
\centering
\includegraphics[width=1.\linewidth]{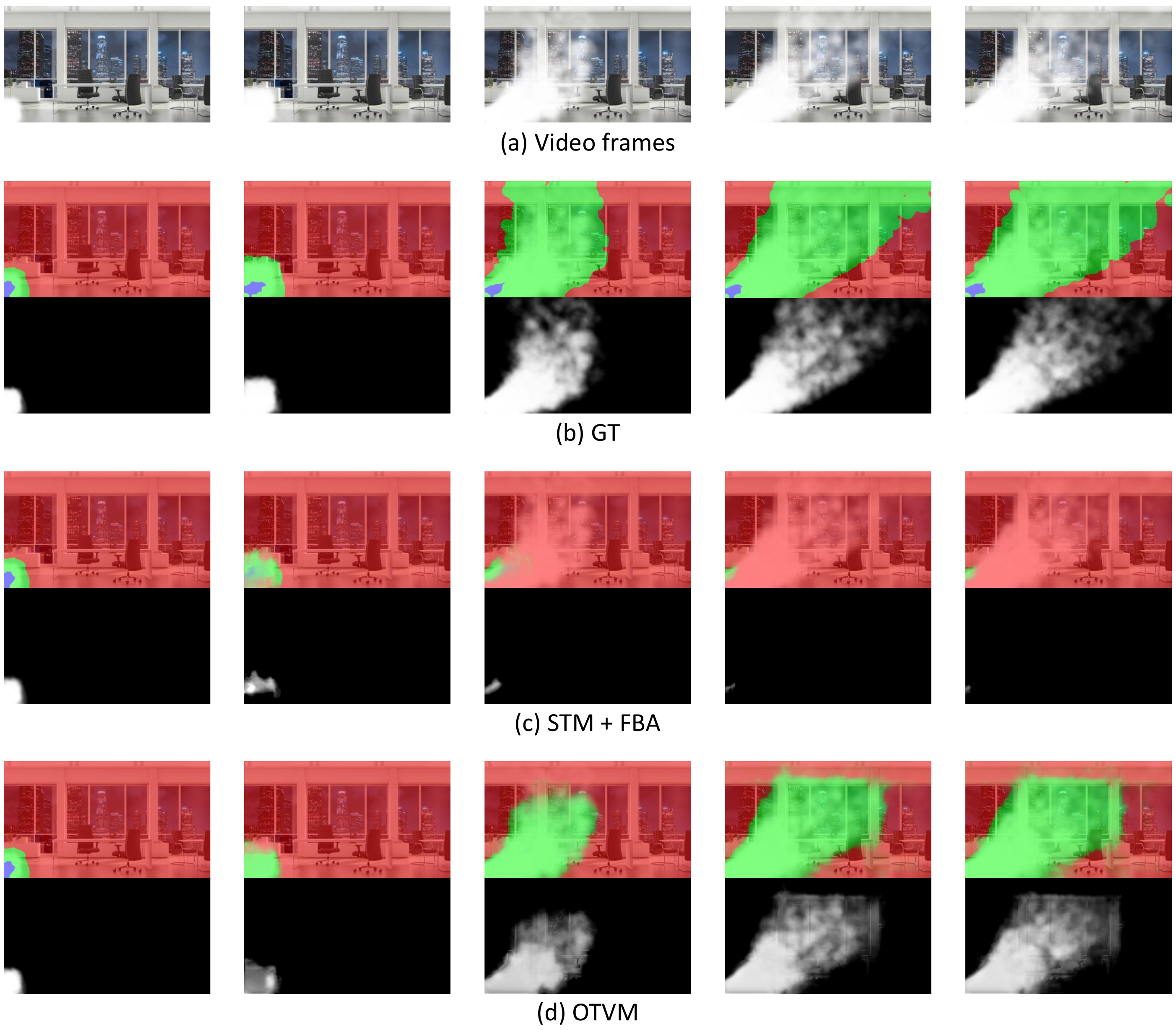}
\caption{
Qualitative results on VideoMatting108 validation set.}
\label{fig:supp_qualitative_results_V108_2}
\end{figure}

\begin{figure}
\centering
\includegraphics[width=1.\linewidth]{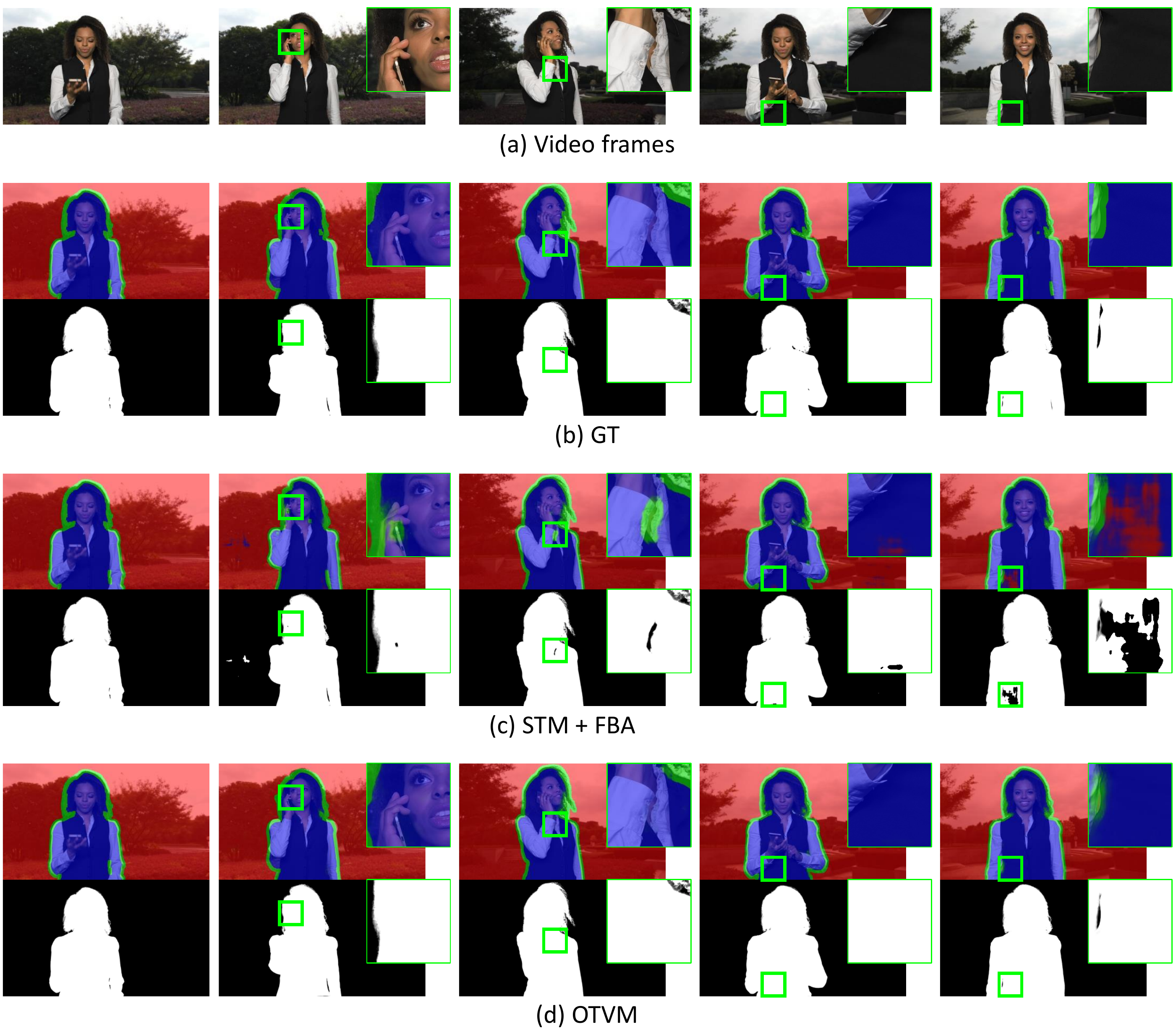}
\caption{
Qualitative results on VideoMatting108 validation set.}
\label{fig:supp_qualitative_results_V108_3}
\end{figure}

\begin{figure}
\centering
\includegraphics[width=1.\linewidth]{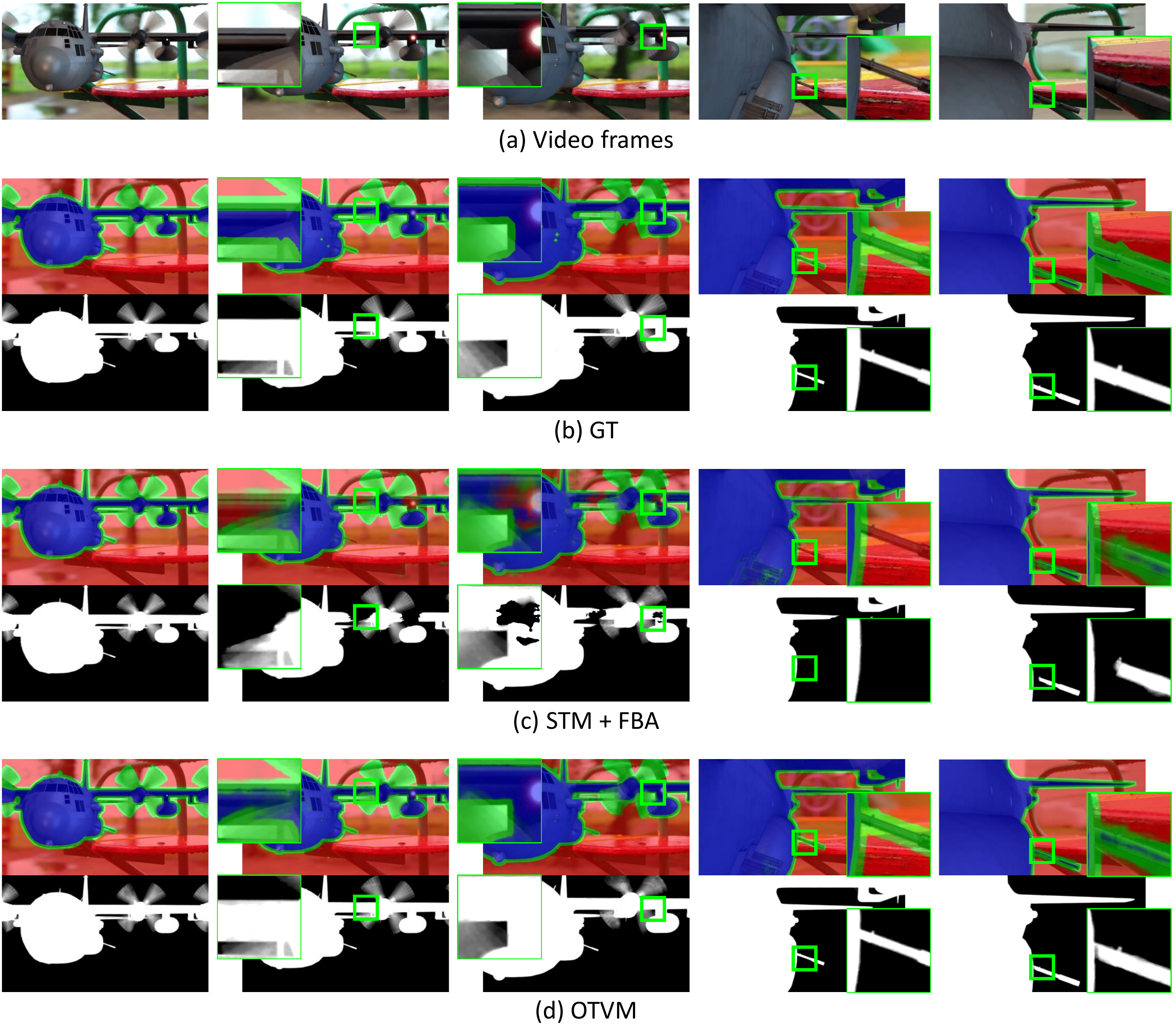}
\caption{
Qualitative results on DVM validation set.}
\label{fig:supp_qualitative_results_DVM_1}
\end{figure}

\begin{figure}
\centering
\includegraphics[width=1.\linewidth]{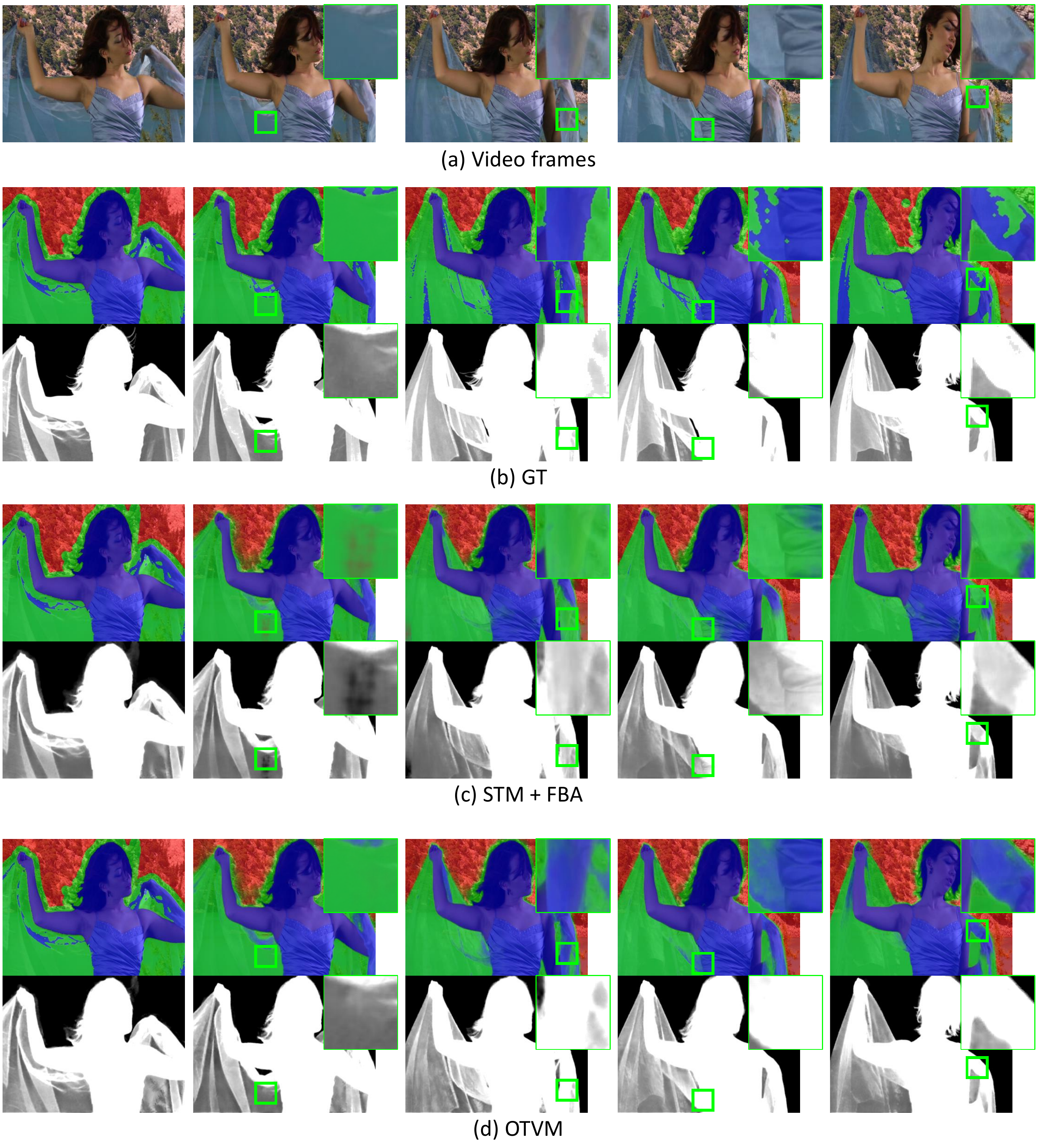}
\caption{
Qualitative results on DVM validation set.}
\label{fig:supp_qualitative_results_DVM_2}
\end{figure}

We present additional qualitative results on real-world videos in \cref{fig:supp_qualitative_results_real_video_1,fig:supp_qualitative_results_real_video_2}, results on VideoMatting108~\cite{zhang2021attention} with medium width trimap in \cref{fig:supp_qualitative_results_V108_1,fig:supp_qualitative_results_V108_2,fig:supp_qualitative_results_V108_3}, and results on DVM~\cite{sun2021deep} in \cref{fig:supp_qualitative_results_DVM_1,fig:supp_qualitative_results_DVM_2}.
We provided user-annotated (or GT) trimap only at the first frame.
For all qualitative results in this section, we further provide full-frame results online: \url{https://youtu.be/qkda4fHSyQE}.

\end{document}